\documentclass{article}

\usepackage{arxiv}

\usepackage[utf8]{inputenc} % allow utf-8 input
\usepackage[T1]{fontenc}    % use 8-bit T1 fonts
\usepackage{hyperref}       % hyperlinks
\usepackage{url}            % simple URL typesetting
\usepackage{booktabs}       % professional-quality tables
\usepackage{amsfonts}       % blackboard math symbols
\usepackage{nicefrac}       % compact symbols for 1/2, etc.
\usepackage{microtype}      % microtypography

\usepackage{graphicx}
\usepackage[numbers]{natbib}
\usepackage{doi}
\usepackage{amsmath}
\usepackage{cleveref}
\usepackage{todonotes}
\usepackage{subcaption}
\usepackage{lscape} 
\usepackage[pagewise]{lineno}%\linenumbers

\newcommand*{\vertbar}{\rule[-1ex]{0.5pt}{2.5ex}}

\title{SINDy-KANs: Sparse identification of non-linear dynamics through Kolmogorov-Arnold networks}

\date{} 					% Or removing it

\author{ 
        \href{https://orcid.org/0000-0002-6411-6198}{\includegraphics[scale=0.06]{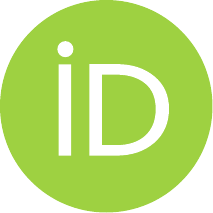}\hspace{1mm}Amanda A. Howard}\\
	Pacific Northwest National Laboratory\\
	Richland, WA 99354 \\
	\texttt{amanda.howard@pnnl.gov} \\
    \And  
            \href{https://orcid.org/0009-0003-3627-5217}{\includegraphics[scale=0.06]{orcid.pdf}\hspace{1mm}Nicholas Zolman}\\
	University of Washington\\
	Seattle, WA 98195 \\
	\texttt{nzolman@uw.edu} \\
    \And
                \href{https://orcid.org/0009-0001-5361-3105}{\includegraphics[scale=0.06]{orcid.pdf}\hspace{1mm}Bruno Jacob}\\
	Pacific Northwest National Laboratory\\
	Richland, WA 99354 \\
	\texttt{bruno.jacob@pnnl.gov} \\
    \And
            \href{https://orcid.org/0000-0002-6565-5118}{\includegraphics[scale=0.06]{orcid.pdf}\hspace{1mm}Steven L. Brunton}\\
	University of Washington\\
	Seattle, WA 98195 \\
	\texttt{sbrunton@uw.edu} \\
    \And
        \href{https://orcid.org/0000-0002-9928-5637}{\includegraphics[scale=0.06]{orcid.pdf}\hspace{1mm}Panos Stinis} \\
	Pacific Northwest National Laboratory\\
	Richland, WA 99354 \\
        \texttt{panagiotis.stinis@pnnl.gov}  
}

\renewcommand{\shorttitle}{SINDy-KANs}

\hypersetup{
}

\begin{document}
\maketitle

\begin{abstract}
Kolmogorov-Arnold networks (KANs) have arisen as a potential way to enhance the interpretability of machine learning. However, solutions learned by KANs are not necessarily interpretable, in the sense of being sparse or parsimonious. Sparse identification of nonlinear dynamics (SINDy) is a complementary approach that allows for learning sparse equations for dynamical systems from data; however, learned equations are limited by the library. In this work, we present SINDy-KANs, which simultaneously train a KAN and a SINDy-like representation to increase interpretability of KAN representations with SINDy applied at the level of each activation function, while maintaining the function compositions possible through deep KANs. We apply our method to a number of symbolic regression tasks, including dynamical systems, to show accurate equation discovery across a range of systems. 
\end{abstract}

% keywords can be removed

\section{Introduction}

Kolmogorov-Arnold networks  (KANs), which use the Kolmogorov-Arnold Theorem as inspiration, have recently been developed as an alternative to multilayer perceptrons (MLPs) \cite{liu2024kan, liu2024kan20}. Unlike MLPs, which use fixed activation functions with trainable weights, KANs use trainable activation functions. This capability has been promoted as a way to develop interpretable machine learning models, by identifying the trained activation functions. Many variations of KANs have quickly become popular; e.g. physics-informed KANs (PIKANs) \cite{shukla2024comprehensive, wang2024kolmogorov, faroughi2025scientific, kiyani2025optimizer}, graph KANs \cite{kiamari2024gkan, decarlo2024kolmogorovarnoldgraphneuralnetworks, bresson2024kagnnskolmogorovarnoldnetworksmeet}, and  deep operator KANs \cite{abueidda2024deepokan}. For physics-informed  training in \cite{shukla2024comprehensive}, modifications of KANs can have similar accuracy to physics-informed neural networks \cite{raissi2019physics}. KANs have been applied to a wide variety of problems, including satellite image classification \cite{cheon2024kolmogorov}, time-series analysis \cite{vaca2024kolmogorov}, and fluid dynamics \cite{toscano2024inferring, kashefi2024kolmogorov, xiong2025j}. For a summary of advances in training KANs, we refer the reader to recent surveys, including \cite{somvanshi2025survey, noorizadegan2025practitioner, toscano2025pinns}.

Much of the previous literature has applied KANs in a way similar to an MLP, without identifying or analyzing the functions through symbolic regression. Many papers discussing interpretability of KANs, such as \cite{baravsin2025exploring}, qualitatively analyze the learned activation functions, but do not connect them to a learned equation. When symbolic regression is performed with KANs, the learned equations are often quite complex, making interpretability difficult (although still certainly easier than weights and biases learned with MLPs, as noted by \cite{panczyk2025opening}). Significant pruning of the KAN during training before the symbolic regression has also been applied to reduce spurious terms in the learned equations, such as in \cite{gao2024toward}. Nevertheless, symbolic regression has been successful with KANs---such as for radio map predictions \cite{liao2025kan}, for predicting the pressure and flow rate of flexible electrohydrodynamic pumps \cite{peng2024predictive},
and even providing interpretable predictions of cancer development \cite{zhong2024interpretable}. However, these approaches can be limited; e.g., \cite{zhong2024interpretable} only used a network with a single layer, so no interaction between the variables is possible beyond a linear combination. In \cite{suaza2025interpretable}, the authors performed symbolic regression for reservoir water temperatures, progressively increasing the number of input variables. When the number of variables was large the accuracy improved through nested nonlinearities, but this reduced the interpretability of the model. Recent work has extended the symbolic regression functionality of KANs by using a large language model to identify the target functions \cite{harvey2025symbolic}. In a different approach, \cite{buhler2025kan} improved symbolic regression with KANs by breaking the problem into smaller decompositions, which allows for iteratively training shallower (single layer) KANs.  

One issue with symbolic regression with KANs is that in \cite{liu2024kan}, the activation functions are identified by comparing with a library of candidate functions. As noted in \cite{bagrow2025softly}, the learned activation functions will not necessarily align with the candidate functions, even if it is known that the candidate functions can be composed to output the target function. This limits the applicability of symbolic regression as implemented in \cite{liu2024kan}, and necessitates work that causes the activation functions to necessarily align with the candidate functions. For example, \cite{bagrow2025softly} uses a dictionary of symbolic and dense terms as the candidate function, with learnable gates that sparsify the representation. Here, we aim to make symbolic regression performed with KANs  more interpretable by directly learning compositions of sparse equation representations thorough a SINDy-like approach. 

The sparse identification of nonlinear dynamics (SINDy) framework \cite{brunton2016discovering} is a widely adopted method for identifying sparse, interpretable models of dynamical systems from data. SINDy has been widely applied, including turbulence modeling \cite{beetham2020formulating, beetham2021sparse}, network modeling \cite{mangan2017inferring}, and model predictive control \cite{kaiser2018sparse}, and has been extended for reinforcement learning \cite{zolman2025sindy},  extremely noisy data \cite{fasel2022ensemble,schaeffer2017sparse, reinbold2020using, messenger2021weak, messenger2021weakGalerikin}, and nonlinear dynamical systems \cite{goyal2022discovery}, among many other extensions. One extension most similar to this work is ADAM-SINDy \cite{viknesh2024adam}, where the SINDy coefficients are determined by the ADAM optimizer \cite{kingma2014adam}. Compared with standard SINDy, ADAM-SINDy allows for parameterized libraries, enabling more customization for the candidate functions.  

In this work we present SINDy-KANs, which combine the sparse function identification of SINDy with the deep learning of KANs. Through KAN layers, SINDy-KANs allow for symbolic regression of function compositions not possible with SINDy. Additionally, SINDy-KANs enforce learning parsimonious, interpretable equations more directly than standard KANs. In a sense, SINDy-KANs can be thought of as a deep version of ADAM-SINDy \cite{viknesh2024adam}, where the SINDy coefficients are directly learned. Both SINDy and KANs at the basic level learn basis function expansions, so combining the methods is especially harmonious. We present the methodology behind SINDy-KANs in Secs. \ref{sec:background-method} and \ref{sec:method} and a series of experiments for both discovery of equations and dynamical systems from data in Sec. \ref{sec:results}. 

\section{Background methodology} \label{sec:background-method}

\subsection{Kolmogorov-Arnold networks }
Kolmogorov-Arnold networks (KANs) \cite{liu2024kan} approximate a multivariate function $f(\mathbf{x})$ by composition and addition of univariate functions. We consider a KAN with $L$ layers and $\{n_j\}_{j=0}^{L}$ nodes per layer (denoted $[n_0, n_1, \ldots, n_L]$). We denote the activation function that connects the $(\ell, i)$ and $(\ell+1, j)$ nodes by $\varphi_{\ell, j, i}$. Each KAN layer takes as univariate input $\mathbf{x}^{(\ell)} = \{x_1^{(\ell)}, \ldots, x_{n_L}^{(\ell)}\}$. The input to the KAN is $\mathbf{x} = \mathbf{x}^{(0)}$ Then, each KAN layer can be expressed as
\begin{equation}
    \mathbf{x}^{(\ell+1)} = \sum_{i=0}^{n_L} \varphi_{\ell, j, i}(x_i^{(\ell)}).
\end{equation}

We denote the output of the KAN by $\mathcal{K}(\mathbf{x})$.% which gives: 
%\begin{equation}
%    \mathcal{K}(\mathbf{x}) = \sum_{i_{L-1}=1}^{n_{L-1}}\varphi_{L-1, i_{L}, i_{L-1}}
%    \left(\sum_{i_{L-2}=1}^{n_{L-2}} \ldots
 %   \left(\sum_{i_{1}=1}^{n_{1}}\varphi_{1, i_2, i_1}
  %  \left(\sum_{i_{0}=1}^{n_{0}}\varphi_{0, i_1, i_0}(x_{i_0})
   %\right)
   % \right) \ldots
    %\right), \label{eq:KAN}
%\end{equation}
The activation functions are represented (on a grid with $g$ points) by a weighted combination of a basis function $b(x)$ and a B-spline, 
\begin{equation}
    \varphi(x) = w_b b(x) + w_s \text{spline}(x) \label{eq:phi_format},
\end{equation}
where 
$$
    b(x) = \frac{x}{1+e^{-x}},
$$
and
$$
    \text{spline}(x) = \sum_i c_i B_i(x).
$$
Here, $B_i(x)$ is a polynomial of degree $k$, typically chosen with $k=3$ or $5$ and fixed as $k=5$ in this work. $c_i, w_b,$ and $w_s$ are trainable parameters if not predetermined by the user. We refer the reader to \cite{liu2024kan} for details on training KANs. 

Data-driven KANs are trained by minimizing the loss function given by the difference between the KAN representation and provided data: 
\begin{equation}
    \mathcal{L}_{\text{KAN}} = ||f(\mathbf{x})-\mathcal{K}(\mathbf{x})||_2^2.
    \label{eq:kan_loss}
\end{equation}

While KANs can represent products of the input functions through the sums of quadratics (e.g., $xy = \frac{1}{4}(x+y)^2-\frac{1}{4}(x-y)^2$), it is simpler and easier to interpret when KANs have multiplication nodes added. Multiple methods have been developed to do so, including MultKANs \cite{liu2024kan20} and LeanKAN \cite{koenig2025leankan}. Both of these methods have advantages and disadvantages in terms of accuracy, trainable parameters, and computational complexity. For this work, we introduce multiplication-enabled KANs where some of the sums of activation functions in each layer are replaced by a product of the activation functions. Specifically, for a KAN layer with $n_j$ nodes, we designate $n_j^M$ of the nodes as multiplication nodes, where the input activation functions for that node are multiplied together. This architecture allows for easier discovery of dynamical systems that contain products of the input variables.

\subsection{Sparse identification of nonlinear dynamics}
Sparse identification of nonlinear dynamics (SINDy) \cite{brunton2016discovering} allows for sparse identification of systems of equations governing dynamical systems. We consider a dynamical system of the form
\begin{equation}
    \frac{d}{dt}\mathbf{x}(t) = \mathbf{f}(\mathbf{x}(t)) \label{eq:dyn_sys}
\end{equation}
for $\mathbf{x}\in \mathbb{R}^n$.
Available data are sampled at times $t_1, t_2, \ldots, t_m$ to create two matrices
\begin{equation}
    \mathbf{X} = \begin{bmatrix}
\mathbf{x}^T(t_1)\\
\mathbf{x}^T(t_2)\\
\vdots\\
\mathbf{x}^T(t_m)\\
\end{bmatrix}
= \begin{bmatrix}
x_1(t_1) & x_2(t_1) & \cdots & x_n(t_1) \\
x_1(t_2) & x_2(t_2) & \cdots & x_n(t_2) \\
\vdots & \vdots & \ddots & \vdots \\
x_1(t_m) & x_2(t_m) & \cdots & x_n(t_m) \\
\end{bmatrix}
\end{equation}
and 
\begin{equation}
     \mathbf{\dot X} = \begin{bmatrix}
\mathbf{\dot x}^T(t_1)\\
\mathbf{\dot x}^T(t_2)\\
\vdots\\
\mathbf{\dot x}^T(t_m)\\
\end{bmatrix}
= \begin{bmatrix}
\dot x_1(t_1) & \dot x_2(t_1) & \cdots & \dot x_n(t_1) \\
\dot x_1(t_2) & \dot x_2(t_2) & \cdots & \dot x_n(t_2) \\
\vdots & \vdots & \ddots & \vdots \\
\dot x_1(t_m) & \dot x_2(t_m) & \cdots & \dot x_n(t_m) \\
\end{bmatrix}.
\end{equation}

A library of candidate nonlinear functions is denoted by $\mathbf{\Theta}(\mathbf{X})$, where each column of $\mathbf{\Theta}(\mathbf{X})$ is given by a candidate function: 
\begin{equation}
    \mathbf{\Theta}(\mathbf{X}) = \begin{bmatrix}
    \vertbar & \vertbar & \vertbar &   &  \vertbar &  \vertbar \\
    \mathbf{1}   & \mathbf{X} & \mathbf{X}^{P_2}    & \cdots & \sin(\mathbf{X}) & \cos(\mathbf{X})   \\
    \vertbar & \vertbar & \vertbar &   &  \vertbar &  \vertbar 
    \end{bmatrix},
\end{equation}
where $\mathbf{X}^{P_2}$ denotes the quadratic
nonlinearities in the state variable $\mathbf{x}$.

The goal of SINDy is to solve a sparse regression problem for vectors of coefficients
$\mathbf{\Xi} = [\mathbf{\xi}_1 \mathbf{\xi}_2 \cdots \mathbf{\xi}_n]$ given by 
\begin{equation}
    \mathbf{\dot X}  = \mathbf{\Theta}(\mathbf{X})\mathbf{\Xi} \label{eq:SINDy}. 
\end{equation}
In practice, obtaining sparse coefficients $\mathbf{\Xi}$ is achieved by minimizing the SINDy loss function
\begin{equation}
        \mathcal{L}_{\text{SINDy}} = \| \mathbf{\dot X}  - \mathbf{\Theta}(\mathbf{X})\mathbf{\Xi}\| + \|\mathbf{\Xi}\|_0,
\end{equation}
where $\|\cdot\|_0$ counts the number of nonzero entries.  This non-convex optimization problem is typically solved using sequentially-thresholded least squares regression, which also makes it possible to enforce known symmetries and other physical constraints on the model coefficients~\cite{Loiseau2017jfm}.

While SINDy and its variants have been very effectively applied to a wide range of applications \cite{kaptanoglu2020physics,callaham2021nonlinear,guan2020sparse,Loiseau2020tcfd,deng2021galerkin,Callaham2022jfm,Callaham2022scienceadvances,zanna2020data,schmelzer2020discovery,beetham2020formulating,beetham2021sparse}, a disadvantage is that the candidate functions must be determined by the user. If the dynamical system consists of complex functions, such as compositions of functions, determining the proper library of candidate functions can be difficult and/or result in a very large library.

\section{SINDy-KANs} \label{sec:method}
In SINDy-KANs, we preform a SINDy-like sparse regression at each KAN activation function as the system is training. In particular, consider an activation function $\varphi_{\ell, j, i}$ with input $\mathbf{x}_i^{(\ell)}$.
We assume available data are sampled at times $t_1, t_2, \ldots, t_m$ to create the matrices: 
\begin{equation}
    \mathbf{X}_i^{(\ell)} = \begin{bmatrix}
x_i^{(\ell)}(t_1), & x_i^{(\ell)}(t_2), & \cdots & x_i^{(\ell)}(t_m) 
\end{bmatrix}.
\end{equation}

 We construct a library of candidate functions  
\begin{equation}
    \mathbf{\Theta}_{\ell, i}(\mathbf{X}_i^{(\ell)}) = \begin{bmatrix}
    \vertbar & \vertbar & \vertbar &   &  \vertbar &  \vertbar \\
    \mathbf{1}   & \mathbf{X}_i^{(\ell)} & \mathbf{X}_i^{(\ell)^{P_2}}    & \cdots & \sin(\mathbf{X}_i^{(\ell)}) & \cos(\mathbf{X}_i^{(\ell)})   \\
    \vertbar & \vertbar & \vertbar &   &  \vertbar &  \vertbar 
    \end{bmatrix}.
\end{equation}
We note that since KANs consist of univariate activation functions, $\mathbf{\Theta}_{\ell,  i}(\mathbf{X}_i^{(\ell)})$ is univariate with respect to the input. We also note that $\mathbf{\Theta}_{\ell, i}(\mathbf{X}_i^{(\ell)})$ does not have to be the same for all $\ell$. Depending on knowledge of the problem, different hidden layers in the SINDy-KAN can be represented by a different set of candidate functions.

Then, we define a sparse vector of coefficients $\mathbf{\xi}_{\ell, j, i} $ and
\begin{equation}
    \mathbf{\varphi}_{\ell, j, i} = 
    \begin{bmatrix}
        \mathbf{\varphi}_{\ell, j, i}(x_i^{(\ell)}(t_1)) \\
        \mathbf{\varphi}_{\ell, j, i}(x_i^{(\ell)}(t_2)) \\
        \vdots \\
        \mathbf{\varphi}_{\ell, j, i}(x_i^{(\ell)}(t_m))
    \end{bmatrix}
\end{equation}
such that 
\begin{equation}
    \mathbf{\varphi}_{\ell, j, i} = \mathbf{\Theta}_{\ell, i}(\mathbf{X}_i^{(\ell)})\mathbf{\xi}_{\ell, j, i} . \label{eq:sindy_one_act}
\end{equation}

\begin{figure}[t]
    \centering    \includegraphics[width=0.9\textwidth]{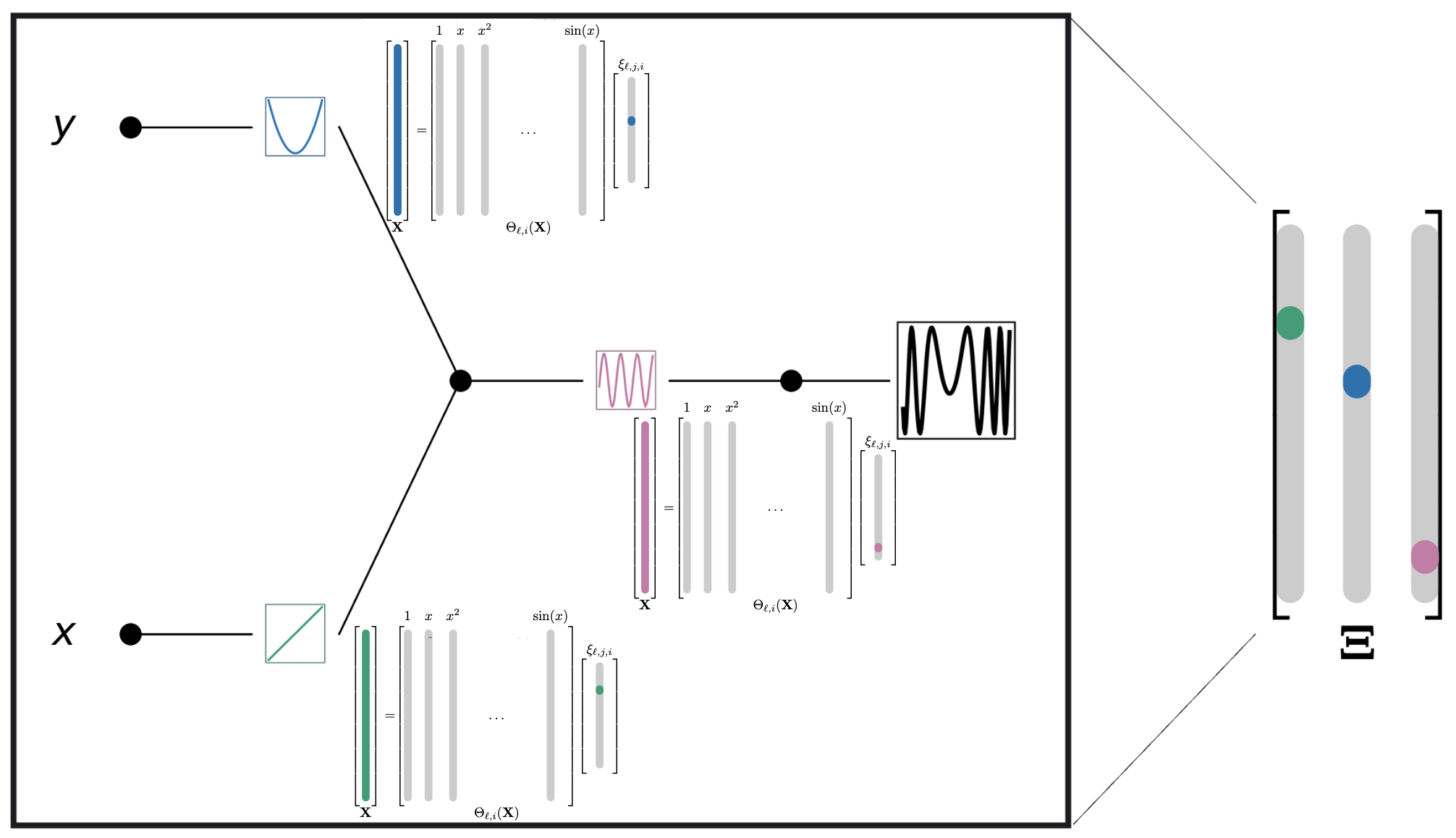}
    \caption{Pictorial representation of a SINDy-KAN for a function $f(x, y) = \sin(x+y^2)$. Each KAN activation function has an associated least squares regression as given in Eq. \ref{eq:sindy_one_act}. The learned coefficients are then combined in $\mathbf{\Xi}$. }
    \label{fig:graphic_sindykan}
\end{figure}

\begin{figure}[t]
    \centering    \includegraphics[width=0.7\textwidth]{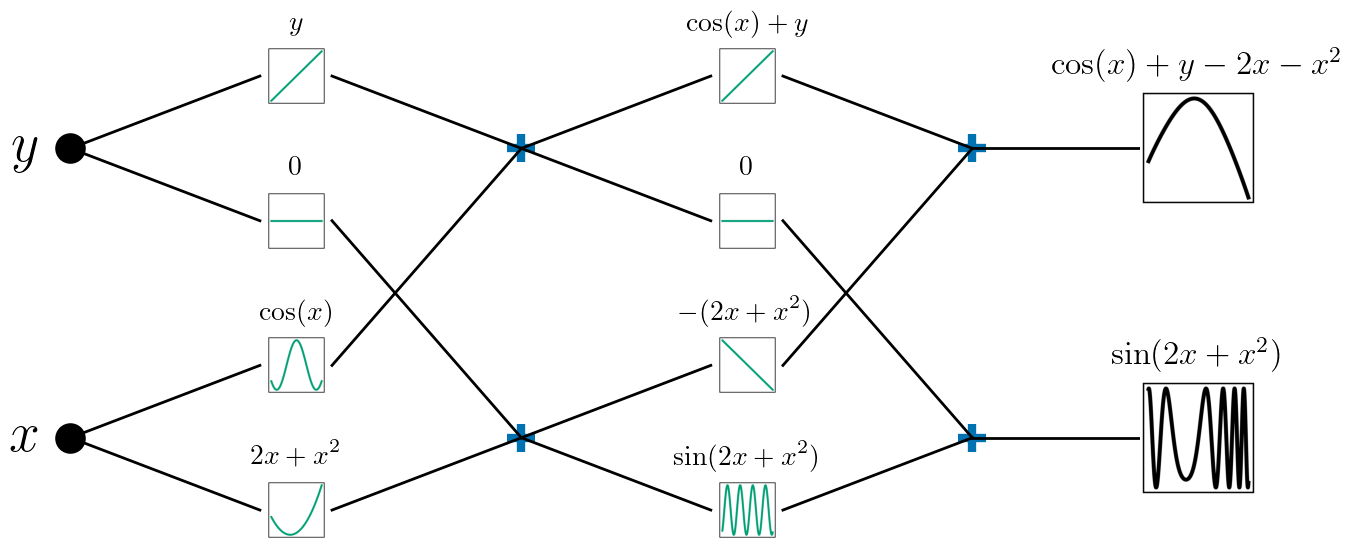}
    \caption{Example of a trained SINDy-KAN for the system of equations $f_1(x, y) = \sin(2x+x^2)$, $f_2(x, y) = \cos(x)+y-2x-x^2$.}
    \label{fig:labeled_sindykan}
\end{figure}

We denote by $\mathbf{\Xi}_{S} = \{\mathbf{\xi}_{1, 1, 1} \mathbf{\xi}_{1, 1, 2} \ldots \mathbf{\xi}_{L, j, i}\}$ to be the learned coefficients.
Once the coefficients $\mathbf{\Xi}_{S}$ are found, the output of the SINDy-KAN can be determined through evaluating the SINDy-KAN representation. % given by 
%\begin{equation}
%    \mathcal{K}_S(\mathbf{x}) = \sum_{i_{L-1}=1}^{n_{L-1}}
 %   \mathbf{\Theta}_{L-1, i_{L}, i_{L-1}}(\mathbf{x})\mathbf{\xi}_{L-1, i_{L}, i_{L-1}}
  %  \left(\sum_{i_{L-2}=1}^{n_{L-2}} \ldots
   % \left(\sum_{i_{1}=1}^{n_{1}}\mathbf{\Theta}_{1, i_2, i_1}(\mathbf{x})\mathbf{\xi}_{1, i_2, i_1}
   % \left(\sum_{i_{0}=1}^{n_{0}} \mathbf{\Theta}_{0, i_1, i_0}(\mathbf{x}_{i_0})\mathbf{\xi}_{0, i_1, i_0} )
    %\right)
    %\right) \ldots
    %\right). \label{eq:SINDy-KAN}
%\end{equation}
 A diagram of a SINDy-KAN is given in Fig. \ref{fig:graphic_sindykan}. A labeled example of a trained SINDy-KAN is given in Fig. \ref{fig:labeled_sindykan}.

A well-trained SINDy-KAN has a few desirable features. The KAN should agree well with the data (the KAN loss defined in Eq. \ref{eq:kan_loss} should be small). Also, the SINDy-KAN representation denoted by $\mathcal{K}_S(\mathbf{x})$ should agree well with the data, giving the error term
\begin{equation}
    \mathcal{L}_{S} = ||f(\mathbf{x})-\mathcal{K}_S(\mathbf{x})||_2^2 \label{eq:sindy_kan_loss}.
\end{equation}
The coefficients 
$\mathbf{\Xi}_{S}$ should be sparse, so $|| \mathbf{\Xi}_{S} ||_1$ is minimized. However, SINDy-KANs are trained using standard neural network optimizers such as ADAM, which struggle with minimizing the $L_1$ norm of $\mathbf{\Xi}_{S}$. Instead, we introduce a shadow matrix $\mathbf{\Lambda}$ that consists of trainable entries that represent a sparse version of $\mathbf{\Xi}_{S}$. $\mathbf{\Lambda}$ is trained to minimize 
$\lambda || \mathbf{\Lambda} ||_1 +  || \mathbf{\Lambda}- \mathbf{\Xi}_{S}||_2^2$. 

We can introduce two additional loss terms. We denote by $\mathcal{K}_\Lambda(\mathbf{x})$ the KAN evaluated with $\Lambda$. Then, we can also consider 
\begin{equation}
    \mathcal{L}_{\Lambda} = ||f(\mathbf{x})-\mathcal{K}_\Lambda(\mathbf{x})||_2^2. \label{eq:lambda_loss}
\end{equation}

The general SINDy-KAN loss function takes the form: 
\begin{equation}
    \mathcal{L} = \lambda_{KAN} \mathcal{L}_{KAN}
        +\lambda_{S} \mathcal{L}_{S}
        +\lambda_{\Lambda} \mathcal{L}_{\Lambda}
        +\lambda_1 || \mathbf{\Lambda} ||_1
        +\lambda_2 || \mathbf{\Lambda}- \mathbf{\Xi}_{S}||_2^2. \label{eq:total_sindy_loss}
\end{equation}
Because $|| \mathbf{\Lambda} ||_1$ should be $\mathcal{O}(1)$, we note that $\lambda_1$ should be on the order of the expected loss $\mathcal{L}_{S}$. For the examples in this work we fix $\lambda_{KAN}=1.0$.

We compare our results with the \texttt{pykan} package, which also does equation discovery as implemented in \cite{liu2024kan}. In \cite{liu2024kan}, each activation function is restricted to have the form $a+bh(cx +d)$ for scalars $a,b,c,$ and $d$ and a function $h$ found from a given library of functions. In particular, this formulation prevents linear combinations of the library functions, which limits the functions the method can identify. The functions are selected at the end of training, instead of during training, so there is no requirement that the learned activation functions should align with the functions in the library, as noted by \cite{bagrow2025softly}. \texttt{pykan} training involves a series of training and sparsification steps that must be managed by the user. In this work, we keep the \texttt{pykan} implementation as close to the implementation in \cite{liu2024kan} as possible. 

%The final method, which we call unconstrained SINDy-KANs, selects the library functions while training but does not enforce sparsity. That is, $\lambda_{\Lambda} = \lambda_1=\lambda_2=0$, and the selected equation is chosen as $\mathcal{K}_S(\mathbf{x})$. Comparing with unconstrained SINDy-KANs can give insight into the necessity of each term in the loss function. 

SINDy-KANs train a standard KAN and simultaneously find
the coefficients $\mathbf{\xi}_{\ell, j, i}$ by solving  Eq. \ref{eq:sindy_one_act} for each activation function using sparse regression. In other words, SINDy-KANs learn the sparse representation \emph{and} the KAN representation simultaneously. %The least squares solves on each activation function return some residual $r_{\ell, j, i}$. Minimizing $r_{\ell, j, i}$ indicates that the KAN representation and the SINDy representation agree well for that activation function. 
Alternatively, one could train a SINDy-KAN to learn the coefficients $\mathbf{\xi}_{\ell, j, i}$ by minimizing Eq. \ref{eq:sindy_kan_loss}. We call this method \emph{direct SINDy-KANs} because the coefficients are learned directly. We discuss direct SINDy-KANs more in Appendix \ref{app:direct}. While direct SINDy-KANs can accurately train, we find that they are less robust than finding the coefficients $\mathbf{\xi}_{\ell, j, i}$ as presented above. They are, however, less expensive to train for large networks, because the computational work at each iteration is significantly lower without solving the sparse regression problems at each KAN node. We leave a further exploration of direct SINDy-KANs for future work.

\section{Experiments} \label{sec:results}
We implemented the SINDy-KAN algorithm in \texttt{jaxKAN} \cite{rigas2025jaxkan}.  \texttt{jaxKAN} offers several advances in KAN training aimed at increasing accuracy for scientific machine learning including advanced initialization, adaptive grid refinement \cite{rigas2025initialization, rigas2024adaptive, rigas2025towards}, and additional KAN basis functions \cite{ss2024chebyshev}. However, in this work we use the B-spline implementation closest to the original KAN formulation in \cite{liu2024kan} to try and provide direct comparisons with existing literature. 

Training parameters for all examples are given in Appendix~\ref{app:params} in~\cref{tab:params_regression} and~\cref{tab:params_ODE}.

\subsection{Symbolic regression} \label{sec:reg_test2}
In this section we focus on learning symbolic equations directly from data for the equation \begin{equation}
    f(x, y) = \cos(x^2+y), 
\end{equation} 
chosen because it contains a composition of functions ($\cos$ and $x^2+y$), something that would be difficult to know to include in a typical library of functions for standard SINDy. KANs are able to handle composition by increasing the depth of the network. We take $(x, y) \in [-2.5, 2.5]\times[-2.5, 2.5]$ and use 1,000 randomly selected points as training data. We take the library of candidate functions $\mathbf{\Theta}_{\ell, j, i}(\mathbf{x})$ as polynomials up to degree two plus sine and cosine, 

\begin{equation}
    \mathbf{\Theta}_{\ell, i} = \begin{bmatrix}
    \vertbar & \vertbar & \vertbar &    \vertbar &  \vertbar \\
    \mathbf{1}   & \mathbf{X}_i^{(\ell)} & \mathbf{X}_i^{(\ell)^{2}}    &  \sin(\mathbf{X}_i^{(\ell)}) & \cos(\mathbf{X}_i^{(\ell)})   \\
    \vertbar & \vertbar & \vertbar &    \vertbar &  \vertbar 
    \end{bmatrix}.
\end{equation}

Hyperparameters used for training are given in Table \ref{tab:params_regression}.

\begin{figure}[t]
   \begin{subfigure}[b]{0.55\textwidth}
    \centering    \includegraphics[width=\textwidth]{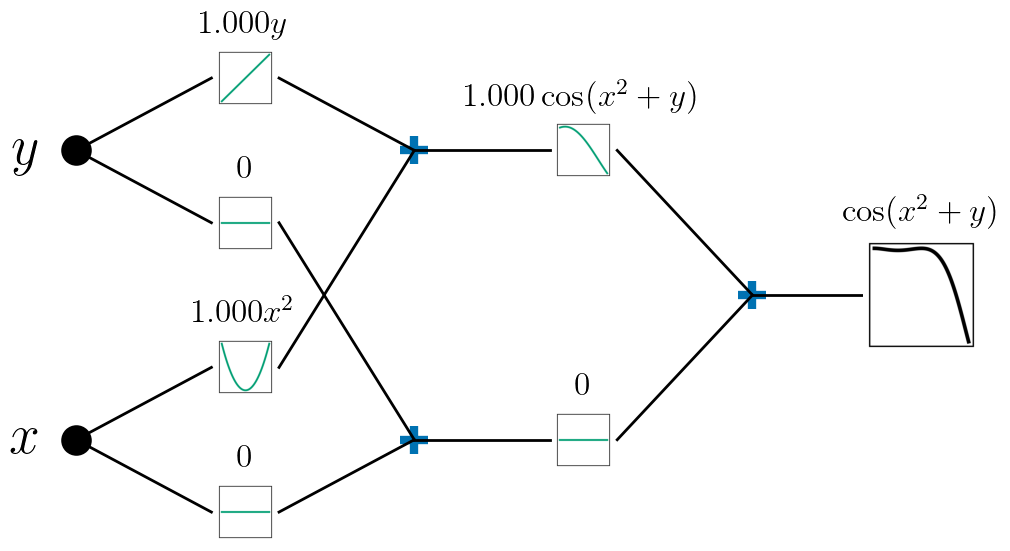}
    \caption{The final trained  SINDy-KAN. }
    \label{fig:Test2_KAN}
\end{subfigure}
   \begin{subfigure}[b]{0.42\textwidth}
    \centering    \includegraphics[width=\textwidth]{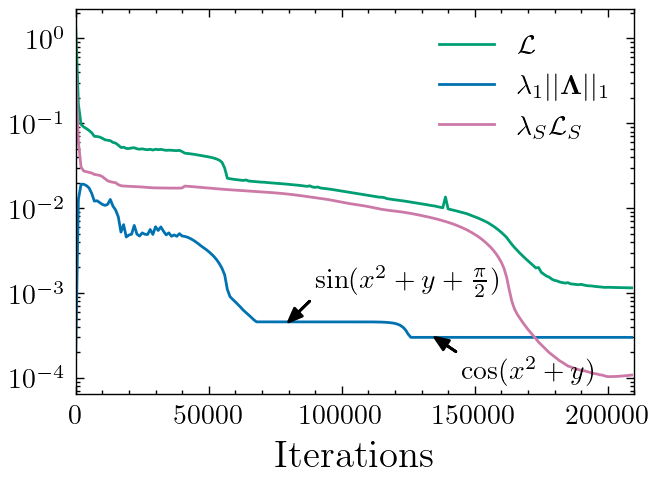}
    \caption{Loss terms $\mathcal{L}$, $\lambda_1 || \mathbf{\Lambda} ||_1$, and $\lambda_S \mathcal{L}_S$.     }
    \label{fig:Test2_KAN_loss}
\end{subfigure}
\caption{Results for Sec. \ref{sec:reg_test2}. (a) The final trained SINDy-KAN correctly learns the target equation. (b) The loss terms show two plateaus for the $L_1$ regularization term $\lambda_1 || \mathbf{\Lambda} ||_1$.}
\end{figure}

The SINDy-KAN learns the correct equation, $$\mathcal{K}_\Lambda=0.9999\cos(1.0000x^2 + 0.9999y),$$ also shown in~\cref{fig:Test2_KAN}. \texttt{pykan} struggles to learn the composition of functions, resulting in 
$$ \mathcal{K}^{pykan}=0.02002x - 0.0275y- 0.5642\sin(1.7946x - 0.1641y + 3.4794). $$
In particular, \texttt{pykan} misses the $x^2$ term, resulting in larger errors overall. 

Examining the loss profiles gives some interesting insight into the SINDy-KAN training. In~\cref{fig:Test2_KAN_loss}, we show the total loss $\mathcal{L}$ given by Eq. \ref{eq:total_sindy_loss} and the $L_1$ regularization term $\lambda_1 || \mathbf{\Lambda} ||_1$. After the initial transient training period, $|| \mathbf{\Lambda} ||_1$ has two plateaus. Examination of the learned equations at the plateaus shows that the first is at $|| \mathbf{\Lambda} ||_1 \approx 3 + \pi/2$, where the SINDy-KAN learns $\sin(x^2 +y + \pi/2)$, which simplifies to the correct equation but is not the sparsest form of the equation. After additional training, the second plateau occurs at $|| \mathbf{\Lambda} ||_1 \approx 3$, representing the equation $\cos(x^2 +y).$

\subsection{Differential equation discovery}\label{sec:ODE}

A common use case for SINDy is identifying dynamical systems of the form given in Eq. \ref{eq:dyn_sys}, where data is given in the form $(t, \mathbf{x})$.  To use SINDy on this data, it is necessary to take numerical derivatives to find $\frac{d\mathbf{x}}{dt}$. Finding this derivative can be difficult, particularly in the presence of noisy data \cite{brunton2016discovering}. Methods used in other work can certainly be used for SINDy-KANs, such finite differences which has  successfully applied to SINDy before \cite{brunton2016discovering}.

\subsubsection{Linear ODE system}\label{sec:ODE_2}
We begin with the simplest class of dynamical systems to test SINDy-KANs.  Even though these equations can be discovered by simpler methods, such as dynamic model discovery, it is useful to confirm that simple systems are also properly identified with the more general SINDy-KAN infrastructure.

We consider the 3D system of equations 
\begin{equation}
    \frac{d}{dt}\begin{bmatrix} x \\ y \\ z\end{bmatrix}=
    \begin{bmatrix}
        -0.1 & 2 & 0\\
        -2 & -0.1 & 0 \\
        0 & 0  & -0.3
    \end{bmatrix}
    \begin{bmatrix} x \\ y \\ z \end{bmatrix}.
    \end{equation}
 We take the library of candidate functions $\mathbf{\Theta}_{\ell, j, i}(\mathbf{x})$ as polynomials up to degree three, 
\begin{equation}
    \mathbf{\Theta}_{\ell, i} = \begin{bmatrix}
    \vertbar & \vertbar & \vertbar &    \vertbar  \\
    \mathbf{1}   & \mathbf{X}_i^{(\ell)} & \mathbf{X}_i^{(\ell)^{2}}    &  \mathbf{X}_i^{(\ell)^{3}}  \\
    \vertbar & \vertbar & \vertbar &    \vertbar  
    \end{bmatrix}.
\end{equation}
The training data is generated with \texttt{pysindy} \cite{desilva2020, Kaptanoglu2022, Kaptanoglu_pysindy}. Data is generated at 200 evenly spaced points in the interval $t\in[0, 10].$

\begin{figure}[t]
    \centering    \includegraphics[width=0.45\textwidth]{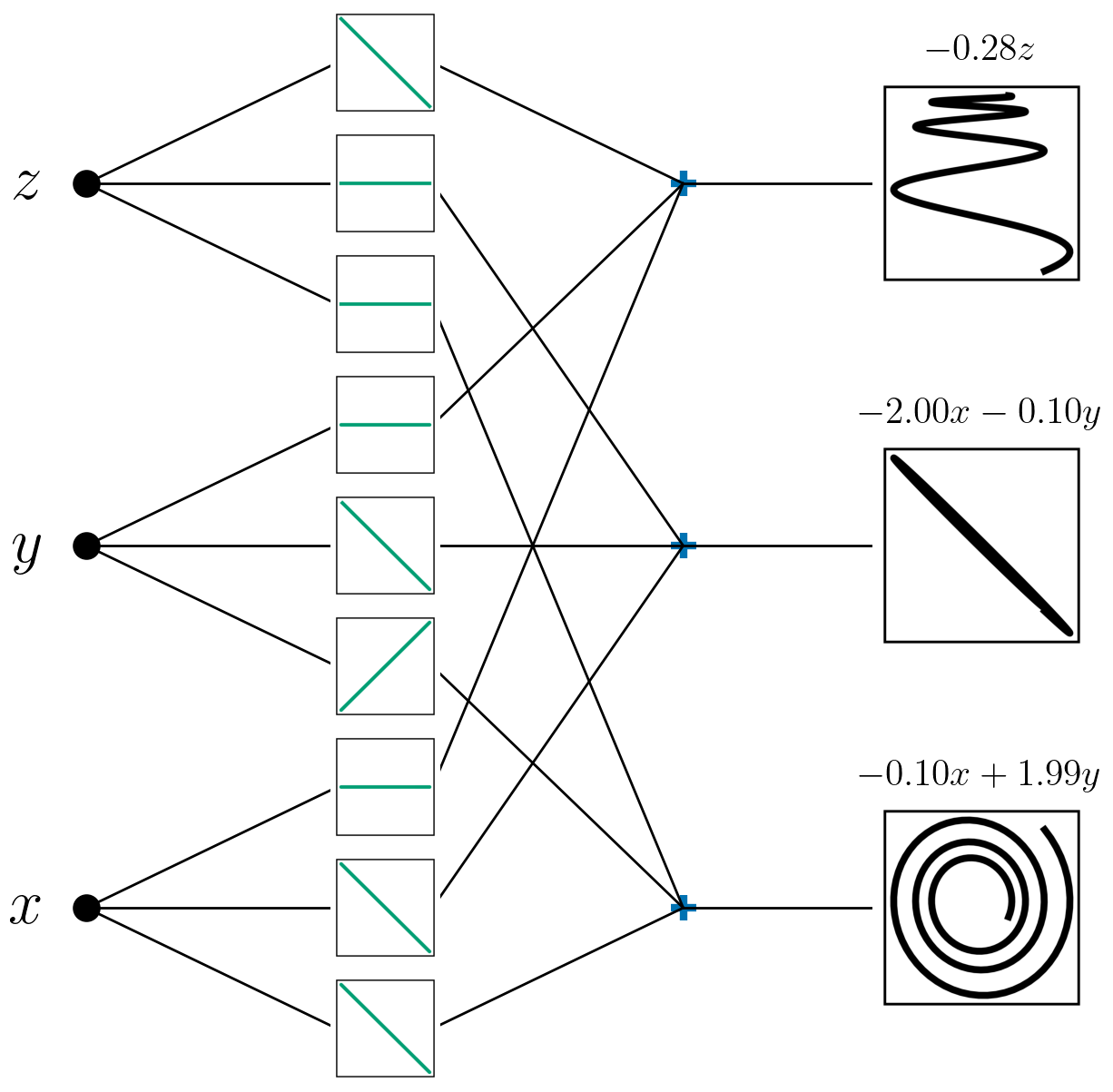}
    \caption{Trained SINDy-KAN for Sec. \ref{sec:ODE_2}.  }
    \label{fig:ODE_2_KAN}
\end{figure}

The SINDy-KAN learns the equation
\begin{equation}
    \frac{d}{dt}\begin{bmatrix} x \\ y \\ z\end{bmatrix}=
    \begin{bmatrix}
        -0.0993 & 1.9946 & 0\\
        -1.9972 & - 0.0995 & 0 \\
        0 & 0  & -0.2811
    \end{bmatrix}
    \begin{bmatrix} x \\ y \\ z \end{bmatrix}.
    \end{equation}
The trained SINDy-KAN is given in Fig. \ref{fig:ODE_2_KAN} and the results are shown in Fig. \ref{fig:ODE_2}. We extend this case to consider training with noise in Appendix ~\ref{app:noise}.

    \begin{figure}[t]
    \centering   
    \includegraphics[width=0.9\textwidth]{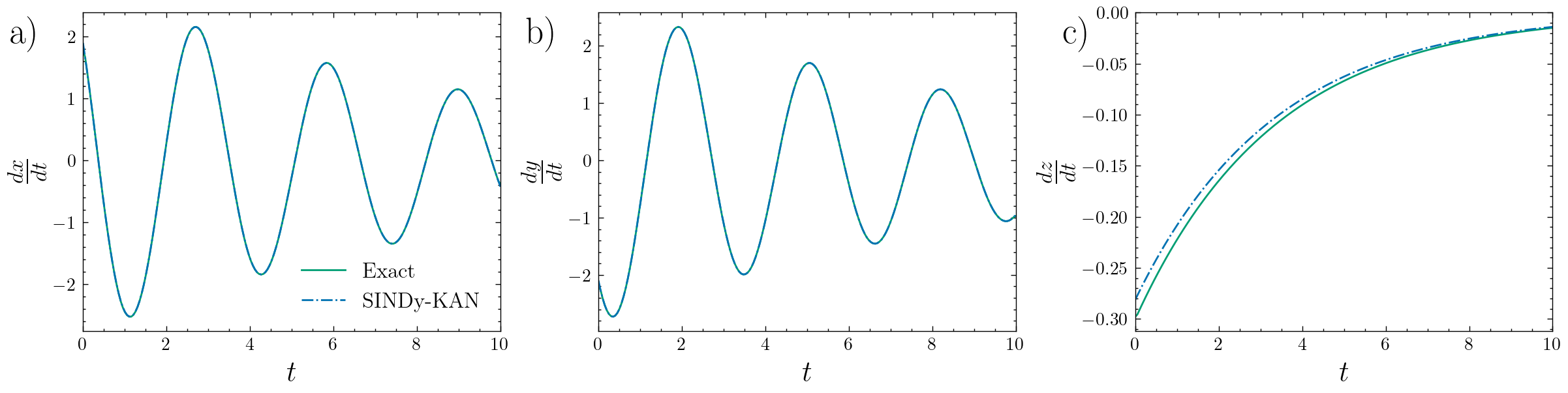}
    \caption{Trained SINDy-KAN for Sec. \ref{sec:ODE_2}. The  SINDy-KAN results agree well with the data. }
    \label{fig:ODE_2}
\end{figure}

\subsubsection{Damped pendulum }\label{sec:ODE_3}
We consider a damped pendulum with equation given by:
\begin{eqnarray}
\frac{dx}{dt} &=& y \label{eq:pendulum_1}\\
\frac{dy}{dt} &=& -0.05y - 9.81 \sin(x)\label{eq:pendulum_2}
\end{eqnarray}
and the library of candidate functions 
\begin{equation}
    \mathbf{\Theta}_{\ell, i} = \begin{bmatrix}
    \vertbar & \vertbar & \vertbar &    \vertbar &  \vertbar \\
    \mathbf{1}   & \mathbf{X}_i^{(\ell)} & \mathbf{X}_i^{(\ell)^{2}}    &  \sin(\mathbf{X}_i^{(\ell)}) & \cos(\mathbf{X}_i^{(\ell)})   \\
    \vertbar & \vertbar & \vertbar &    \vertbar &  \vertbar 
    \end{bmatrix}.
\end{equation}
We can use this test to examine the performance of SINDy-KANs when the architecture is varied by considering SINDy-KANs with one and two hidden activation function layers. The training data is found by numerically solving Eqs. \ref{eq:pendulum_1} and \ref{eq:pendulum_2} at 1000 points in the time interval $t\in[0, 10].$ 

\begin{figure}[t]
       \begin{subfigure}[b]{0.4\textwidth}
    \centering    \includegraphics[width=\textwidth]{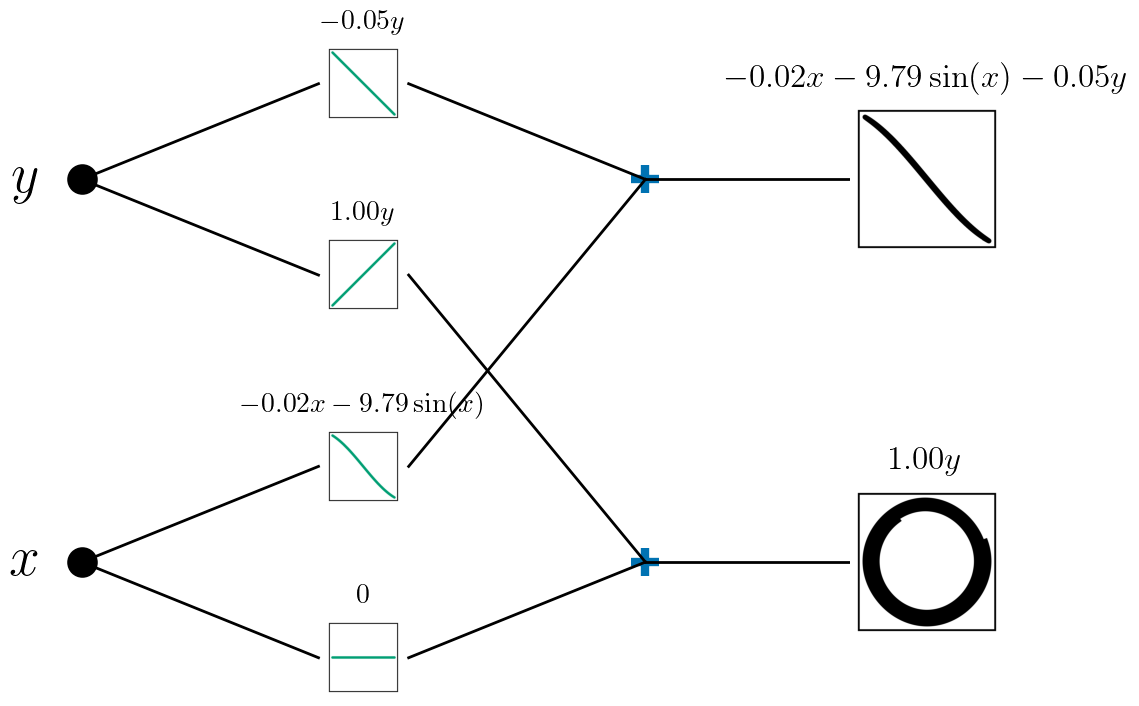}    
    \caption{Trained SINDy-KAN with one hidden layer. }\label{fig:Pendulum_1a}
    \end{subfigure}
           \begin{subfigure}[b]{0.5\textwidth}
    \centering    \includegraphics[width=\textwidth]{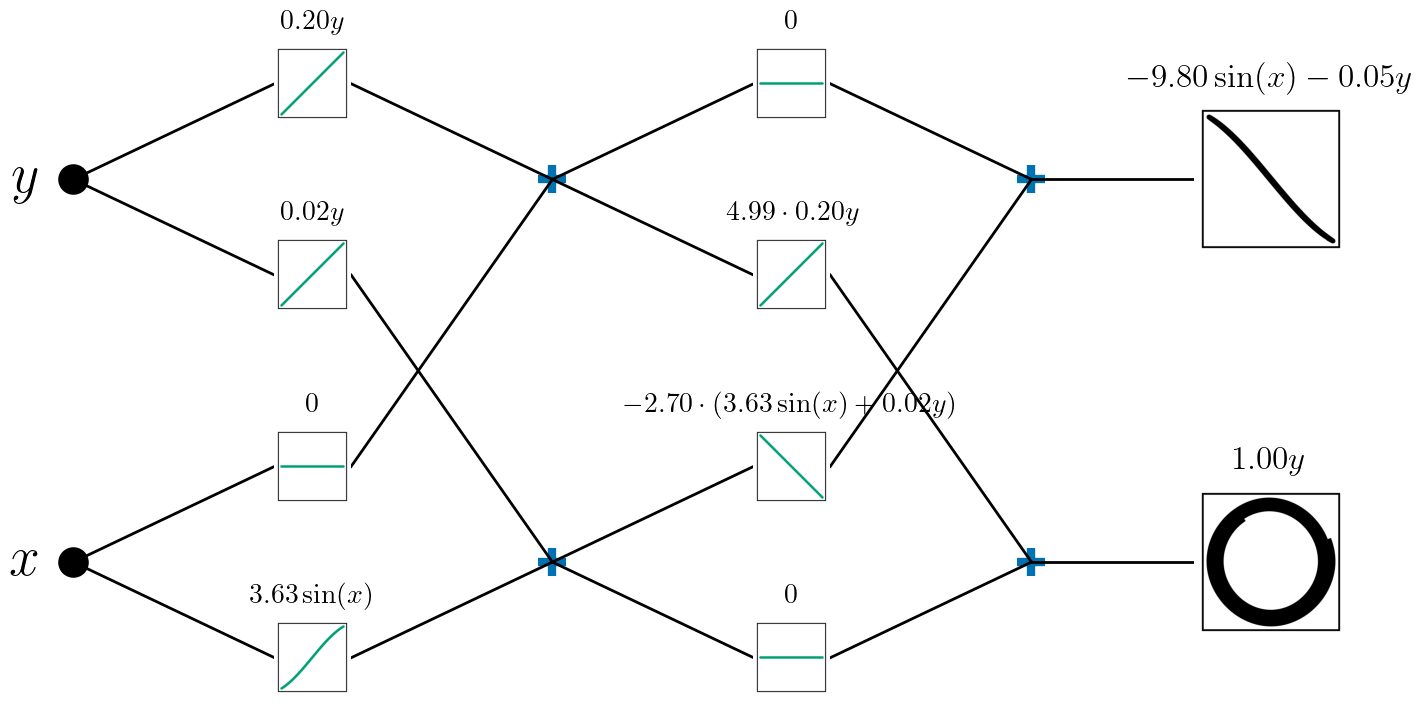}
    \caption{Trained SINDy-KAN with two hidden layers. }\label{fig:Pendulum_1b}
    \end{subfigure}
        \caption{Trained SINDy-KANs for the pendulum  in Sec. \ref{sec:ODE_3} with one (a) and two (b) hidden layers. }
\end{figure}

\begin{figure}[t]
    \centering  
    \includegraphics[width=\textwidth]{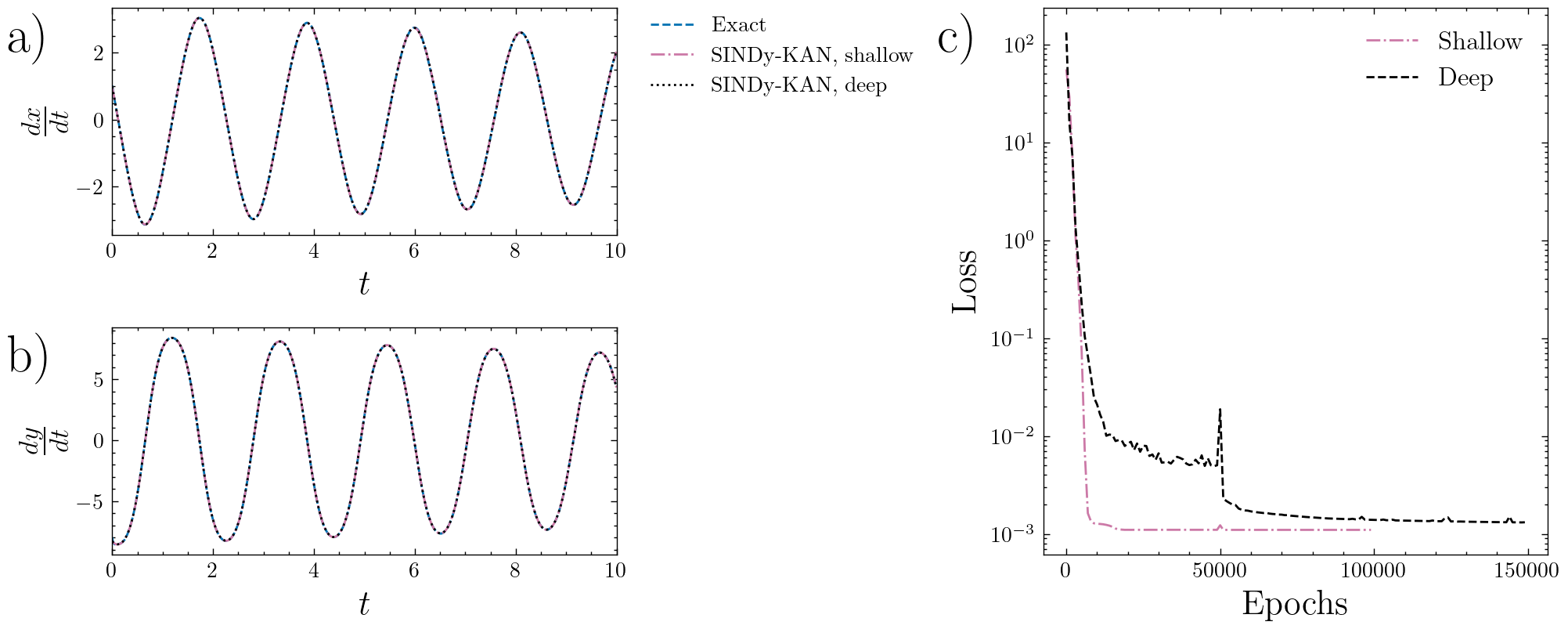}
    \caption{SINDy-KAN results for Sec. \ref{sec:ODE_3}. (a-b) Results for the SINDy-KANs, compared with the numerical derivatives. (c) Loss for the SINDy-KAN. }
    \label{fig:Pendulum_2}
\end{figure}

The SINDy-KAN with one hidden layer learns
\begin{eqnarray*}
\frac{dx}{dt} &=& 0.9998 y \\
\frac{dy}{dt} &=& - 9.7877\sin(x)-0.0499y-0.0185x.
\end{eqnarray*} 
The SINDy-KAN with two hidden layers learns
\begin{eqnarray*}
\frac{dx}{dt} &=& 0.9997 y \\
\frac{dy}{dt} &=& - 9.7975\sin(x)-0.0499y.
\end{eqnarray*}
Plots of the trained SINDy-KAN are shown in~\cref{fig:Pendulum_1a}, \cref{fig:Pendulum_1b} and \cref{fig:Pendulum_2}. The deeper SINDy-KAN is slightly more accurate due to the additional flexibility of the extra hidden layer. However, deeper KANs can also be harder to train in general \cite{Bagrow_2025}, so there is a trade off between the additional flexibility, training time, and complexity of training. 

\subsubsection{ABC flow}\label{sec:ODE_5}
When dynamics of the problem are known they can be incorporated into the SINDy-KAN candidate functions to improve training. For an example, we consider the ABC flow example inspired by \cite{viknesh2024adam}, given by
\begin{eqnarray}
\frac{dx}{dt} &=& A\sin(w_1 z) + C\cos(w_2 y) \label{eq:ABC-1} \\
\frac{dy}{dt} &=& B\sin(w_3 x) + A\cos(w_4 z)\label{eq:ABC-2} \\
\frac{dz}{dt} &=& C\sin(w_5 y) + B\cos(w_6 x) \label{eq:ABC-3} 
\end{eqnarray}
for $A=2$, $B=3$, $C=1$, and     $w_1 =  \pi/4.0$, $w_2 =  \pi/3.0$, $w_3 =  \pi/2.0$, $w_4 =  \pi/5.0$, $w_5 =  \pi/4.5$, $w_6 =  \pi/2.8$.
We use this equation to highlight the selection of different candidate functions for each hidden layer in the KAN. If we know the equations are of the form $\sum_{i=1}^3 a_i\sin(w_i x_i) + b_i\cos(v_i x_i)$ for $i\in\{1, 2, 3\}$ and unknown coefficients $a_i, b_i, w_i$ and $v_i,$ we can consider a two layer KAN, with the candidate functions for the first layer $\{x\}$ and the candidate functions for the second layer $\{\cos(x), \sin(x)\}$. By keeping the first layer as a linear basis to learn the right coordinates/scaling, then feeding them into the nonlinear terms, we restrict the training space to allow for better training while capturing the dynamics of the problem. We emphasize that the frequencies $w_i$ are learned by the SINDy-KAN instead of prescribed by the user. The training data are found by numerically solving the system of equations at 20000 points in the time interval $t\in[0, 20]$. 

Eqns. \ref{eq:ABC-1}-\ref{eq:ABC-3} give approximately:
\begin{eqnarray*}
\frac{dx}{dt} &=& 2\sin(0.7854 z) +1\cos(1.0472 y) \\
\frac{dy}{dt} &=& 3\sin(1.5708 x) + 2\cos(0.6283 z) \\
\frac{dz}{dt} &=& 1\sin(0.6981 y) + 3\cos(1.1210 x) .
\end{eqnarray*}

With the SINDy-KAN, we learn:
\begin{eqnarray*}
\frac{dx}{dt} &=& 2.0072\sin(0.7849z) + 1.0000\cos(1.0470 y) \\
\frac{dy}{dt}&=& 2.9990\sin(1.5708 x) + 1.9926\cos(0.6278 z) \\
\frac{dz}{dt} &=& 0.9996\sin(0.6976 y) + 2.9983\cos(1.2221 x) .
\end{eqnarray*}

\begin{figure}[t]
    \centering   
    \includegraphics[width=0.7\textwidth]{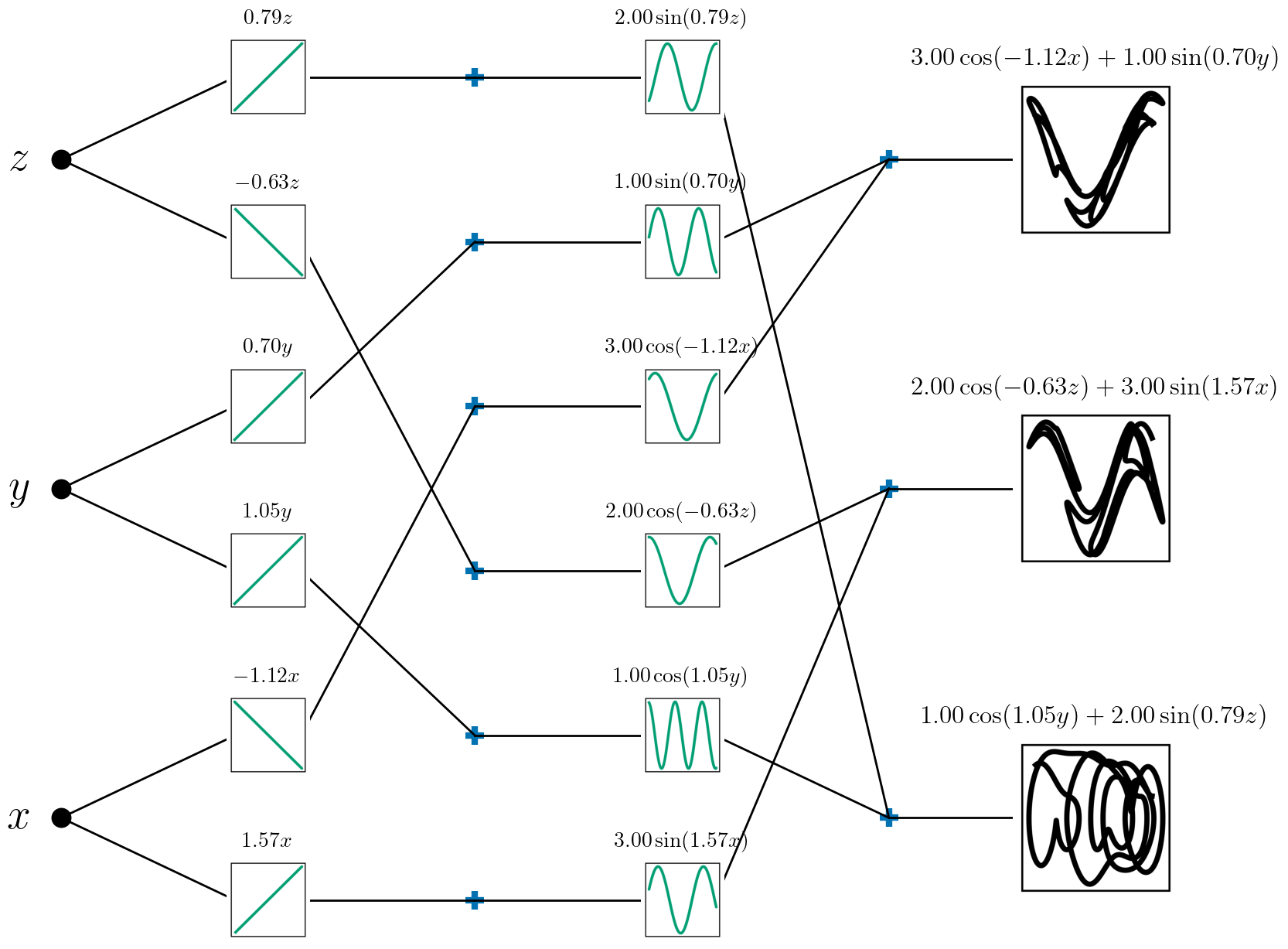}
    \caption{The SINDy-KAN for ABC flow problem in Sec. \ref{sec:ODE_5}. For clarity, only the non-zero activation functions are shown.}
    \label{fig:ABC2}
\end{figure}

\begin{figure}[t]
    \centering   
    \includegraphics[width=0.7\textwidth]{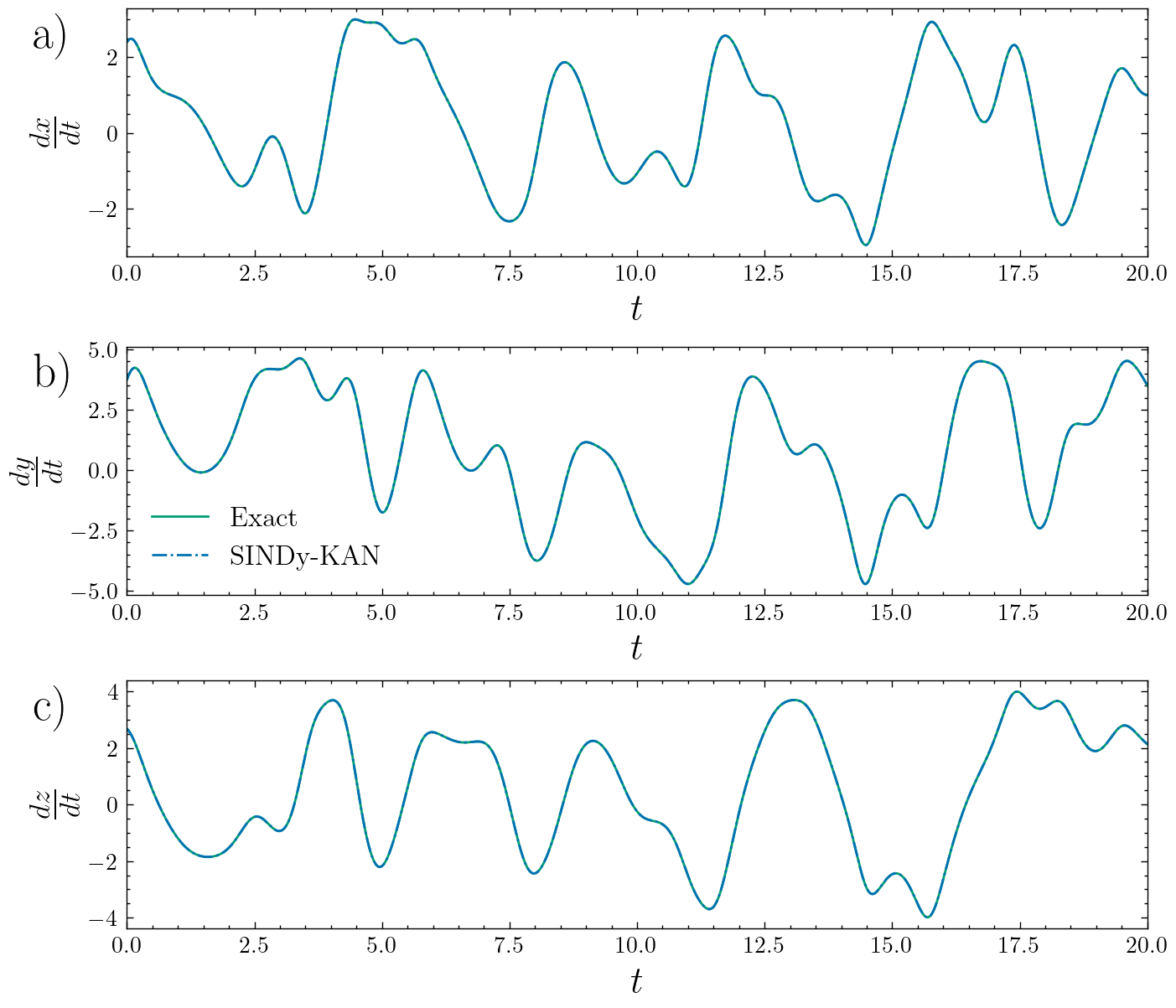}
    \caption{The SINDy-KAN learns the dynamics of the ABC flow problem in Sec. \ref{sec:ODE_5} well. }
    \label{fig:ABC1}
\end{figure}

The trained SINDy-KAN is shown in Fig. \ref{fig:ABC2} and the results from the  SINDy-KAN are shown in Fig. \ref{fig:ABC1}. As in \cite{viknesh2024adam}, SINDy-KANs are able to directly learn the frequencies accurately. This presents a strong advantage over standard SINDy, where the frequencies would need to be predetermined to be included in the library of candidate functions. 

\subsubsection{Lorenz}\label{sec:ODE_6}
We next consider the nonlinear Lorenz system 
\begin{align}
    \frac{dx}{dt} &= \sigma(y-x)\label{eq:l1} \\ 
    \frac{dy}{dt} &= x(\rho-z)-y \label{eq:l2} \\ 
    \frac{dz}{dt} &= xy-\beta z \label{eq:l3}
\end{align}
for     $\rho = 28.0$, 
    $\sigma = 10.0$, and 
    $\beta = 8.0 / 3.0.\approx 2.6667 .$
This example highlights the use of multiplication SINDy-KANs. In particular, we use a KAN with one hidden layer with five nodes, and take two of the nodes as multiplication nodes ($n_1^m = 2$). We take the library of candidate functions for the activation functions as polynomials up to degree one, 
\begin{equation}
    \mathbf{\Theta}_{\ell, i} = \begin{bmatrix}
    \vertbar & \vertbar  \\
    \mathbf{1}   & \mathbf{X}_i^{(\ell)} \\
    \vertbar & \vertbar 
    \end{bmatrix}.
\end{equation}

With the multiplication nodes, the multiplication KAN can represent second degree polynomials after the first layer, even though each activation function is univariate. In contrast, a standard KAN would take two hidden layers to represent second degree polynomials. 

We train the SINDy-KAN using data generated with the initial condition $(-8, 8, 27).$ The training data is found by numerically solving the system of equations at 5000 points in the time interval $t\in[0, 10]$.

\begin{figure}[t]
    \centering   
    \includegraphics[width=0.8\textwidth]{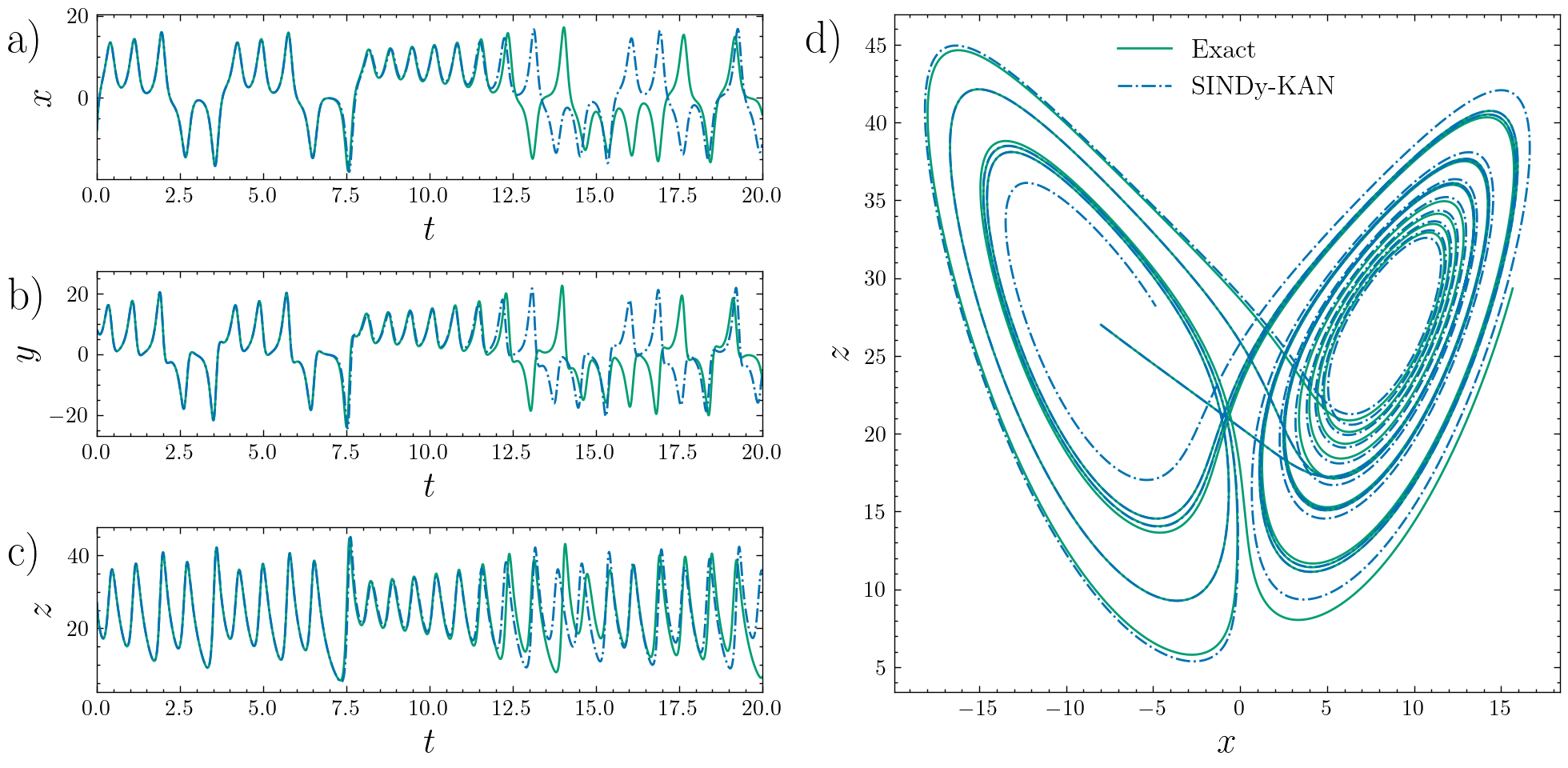}
    \caption{The Lorenz system in Eqs. \ref{eq:l1}--\ref{eq:l3} and the system learned by the SINDy-KAN in Eqs. \ref{eq:sl1}--\ref{eq:sl3} for Sec. \ref{sec:ODE_6}. Both the true system and the learned system are evaluated numerically. We note that only data from $t\in[0, 10]$ is used for training.}
    \label{fig:Lorenz1}
\end{figure}

\begin{figure}[t]
    \centering   
    \includegraphics[width=0.8\textwidth]{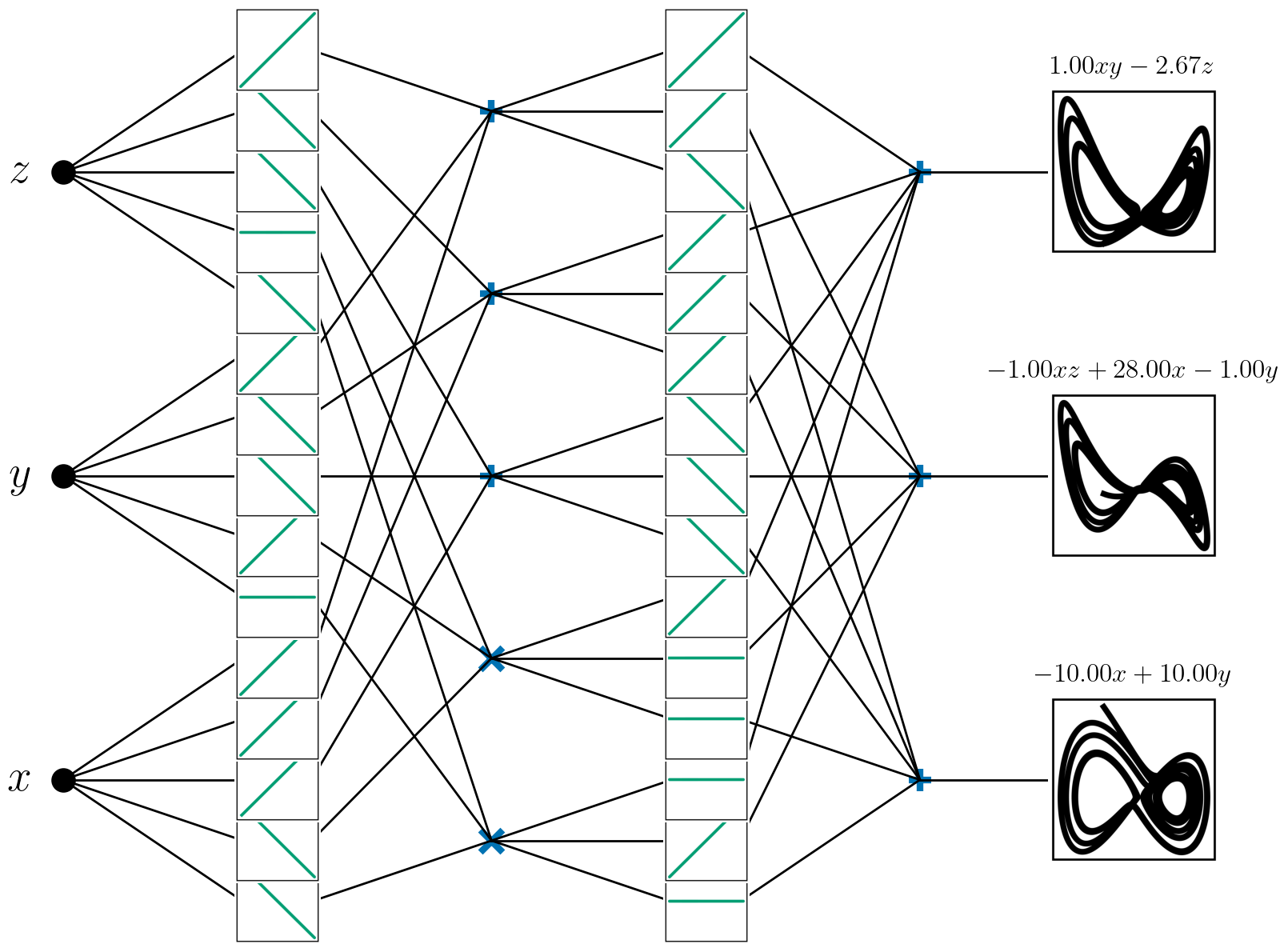}
    \caption{The trained SINDy-KAN for the Lorenz system in Sec. \ref{sec:ODE_6}. Multiplication nodes are denoted by ``$\times$'' and addition nodes are denoted by ``$+$''. }
    \label{fig:Lorenz2}
\end{figure}

The SINDy-KAN learns: 
\begin{align}
    \frac{dx}{dt} &= -10.00001x + 9.99999y \label{eq:sl1} \\
    \frac{dy}{dt}&= x(28.00002-1.00000z)-1.00000y \label{eq:sl2} \\
    \frac{dz}{dt} &= 1.00000xy-2.66666 z.\label{eq:sl3} 
\end{align}
The results are plotted in Fig. \ref{fig:Lorenz1} and the trained KAN is shown in Fig. \ref{fig:Lorenz2}.

\subsubsection{Kuramoto oscillator}\label{sec:ODE_7}
The Kuramoto oscillator is defined by 
\begin{align}
    \frac{d\theta_i}{dt} = \omega_i + \frac{1}{N}\sum_{j=1}^N K_{ij} \sin (\theta_j - \theta_i), \; i = 1, \dots N.
\end{align}
We consider $N = 3$ and $\omega_1 = 16$, $\omega_2 = 5$, and $\omega_3 = 11$ with $K_{ij} = 1$ for all $i, j.$ We consider a two layer KAN, with the candidate functions for the first layer $\{x\}$ and the candidate functions for the second layer $\{1, \sin(x)\}$. The training data is generated with \cite{Kuramoto_rep} at 5000 points for $t\in[0, 5].$

\begin{figure}[t]
    \centering   
    \includegraphics[width=0.8\textwidth]{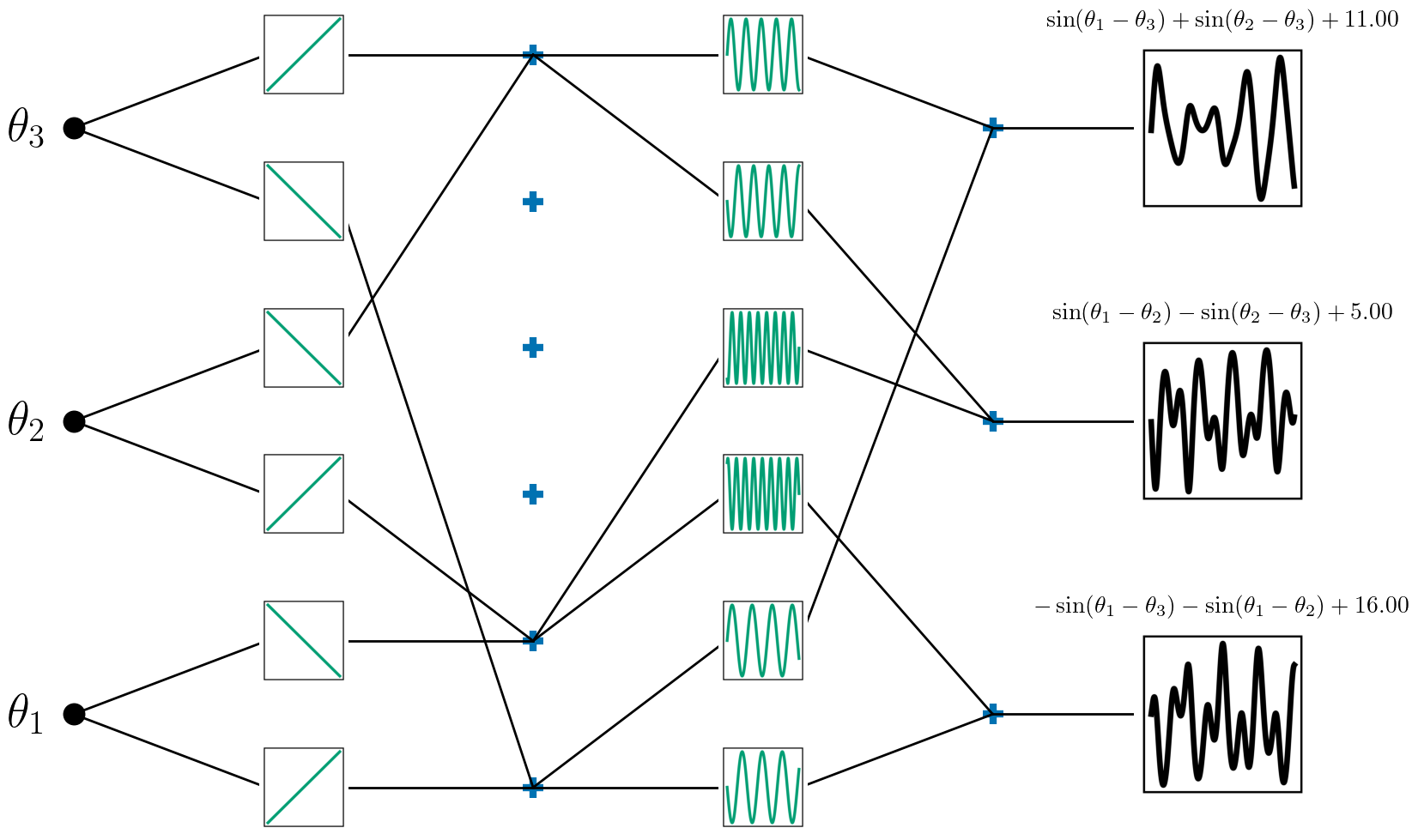}
    \caption{The trained SINDy-KAN for the Kuramoto system in Sec. \ref{sec:ODE_7}. Multiplication nodes are denoted by ``$\times$'' and addition nodes are denoted by ``$+$''. For clarity, only non-zero nodes are shown. }
    \label{fig:Kuramoto}
\end{figure}

The SINDy-KAN learns 
\begin{align}
    \frac{d\theta_1}{dt}  &= -1.0002 \sin(0.9988 \theta_1 - 0.9985 \theta_3) - 1.0001 \sin(0.9994\theta_1 - 0.9981\theta_2) + 15.9995 \\
    \frac{d\theta_2}{dt}  &= 1.0000 \sin(0.9994 \theta_1 - 0.9981 \theta_2) - 1.0002 \sin(0.9988\theta_2 - 0.9990\theta_3) + 5.0002 \\
    \frac{d\theta_3}{dt}  &= 1.0001 \sin(0.9988 \theta_1 - 0.9985 \theta_3) + 1.0002 \sin(0.9988\theta_2 - 0.9990\theta_3) + 11.0003 .
\end{align}
Examining the KAN structure in Fig. \ref{fig:Kuramoto}, the KAN uses only three of the six nodes in the hidden layer (the other nodes are zero.) This minimizes the $\ell_1$ penalty $\lambda_1 || \mathbf{\Lambda} ||_1$ in the loss function by reusing the $\theta_i - \theta_j$ terms in the final equations by recognizing the symmetry that $\sin(x) = -\sin(x).$

\section{Conclusions}
We have shown SINDy-KANs train well and accurately discover equations for a variety of datasets, including chaotic dynamical systems. SINDy-KANs can learn compositions of functions that are difficult to learn with SINDy, where the exact needed composition may not be included in the function library. At the same time, SINDy-KANs more robustly learn parsimonious representations than standard KANs. This combination of function compositions and sparse function representations builds on the strengths of both KANs and SINDy in a harmonious way to offer increased symbolic regression. More broadly, SINDy-KANs contribute to the growing effort to develop interpretable and trustworthy machine learning methods for scientific discovery. By enforcing parsimony at the level of each activation function while retaining the expressiveness of deep networks, SINDy-KANs offer a principled framework for learning human-readable equations from data. 

 Despite these strengths, several limitations remain. As the number of input variables grows, the interpretability of the discovered equations may diminish, a challenge shared with standard KAN-based symbolic regression. Additionally, discovering dynamical systems from data requires numerical differentiation, which can be sensitive to noise. While methods such as finite differences have been successfully applied in SINDy, further work is needed to evaluate the robustness of SINDy-KANs under noisy conditions.

 The examples in this work are all trained on a single Apple M3 Max Macbook Pro, generally in less than a few minutes per problem. Due to not needing to complete the sparse regression at each activation function for each iteration, the direct SINDy-KANs presented in Appendix~\ref{app:direct} train significantly faster. However, SINDy-KANs do not represent a large computational burden. Due to the comparable accuracy between the two approaches, we suggest the faster training times are an argument to use direct SINDy-KANs when possible. However, direct SINDy-KANs assume that each activation function can be represented well by linear combinations of the library functions. In future work, we will consider cases where some activation functions can be kept as their B-spline representations, while others are fixed by their symbolic forms based on the error while training. This will allow for accuracy when the set of library functions is not sufficient to capture the system dynamics. For this approach, SINDy-KANs are necessary.

One consideration when training SINDy-KANs is the network architecture. The depth of the network is the number of function compositions that can occur, and it is difficult to know the necessary number of layers in advance. While SINDy-KANs can learn the identity function for additional unneeded layers, multi-exit KANs could be combined with SINDy-KANs to simultaneously learn the best network architecture and symbolic representation \cite{Bagrow_2025}. Beyond those considered in this work, SINDy-KANs could be combined with many other KAN training schemes to increase accuracy \cite{noorizadegan2025practitioner}, including domain decomposition if different equations are expected in different parts of the domain \cite{howard2025finitebasiskolmogorovarnoldnetworks}, adaptive weighting schemes \cite{anagnostopoulos2023residualbasedattentionconnectioninformation}, automatic grid updates \cite{moody2025automatic}, or improved optimizers \cite{kiyani2025optimizer}. Improved training of the underlying KAN improves the accuracy of the learned equations.

\section{Acknowledgments}
     This project was completed with support from the U.S. Department of Energy, Advanced Scientific Computing Research program, under the Scalable, Efficient and Accelerated Causal Reasoning Operators, Graphs and Spikes for Earth and Embedded Systems (SEA-CROGS) project (Project No. 80278). The computational work was performed using PNNL Institutional Computing at Pacific Northwest National Laboratory. Pacific Northwest National Laboratory (PNNL) is a multi-program national laboratory operated for the U.S. Department of Energy (DOE) by Battelle Memorial Institute under Contract No. DE-AC05-76RL01830. 
     SLB and NZ acknowledge funding support from the National Science Foundation AI Institute in Dynamic Systems (grant number 2112085).

\section{Data and code availability}
The examples in this work were developed using the \texttt{jaxKAN} package \cite{Rigas_jaxKAN_A_JAX-based_2024, Rigas2025}. Results were compared with symbolic regression using \texttt{pykan} \cite{liu2024kan}. Postprocessing of the learned equations was performed with \texttt{sympy} \cite{10.7717/peerj-cs.103}. 

All code and data will be released upon publication. %Meanwhile, we have released code and Google Colab tutorials for SINDy-KANs in Neuromancer~\cite{Neuromancer2023} at \href{https://github.com/pnnl/neuromancer/examples/KANs/p3_sindy_kan_symbolic_regression.ipynb}{https://github.com/pnnl/neuromancer/examples/KANs/p3\_sindy\_kan\_symbolic\_regression.ipynb} for the reader to explore the ideas implemented in this work.

\bibliographystyle{unsrt}
\bibliography{references}  %%% Uncomment this line and comment out the ``thebibliography'' section below to use the external .bib file (using bibtex) .

\clearpage
\appendix

\section{Direct SINDy-KANs} \label{app:direct}

In direct SINDy-KANs, the coefficients $\mathbf{\Xi}_{S}$ are taken as trainable parameters and learned directly (the underlying KAN is not trained.) Direct SINDy-KANs can be thought of as a deep version of ADAM-SINDy \cite{viknesh2024adam}, where instead of learning one set of coefficients, the coefficients for each activation function in a KAN are learned through the ADAM optimizer. Alternatively, direct SINDy-KANs are directly equivalent to Softly Symbolified Kolmogorov-Arnold Networks (S2KANs), with a different method of enforcing sparsity. 

In direct SINDy-KANs,~\cref{eq:total_sindy_loss} is modified to
\begin{equation}
    \mathcal{L}_{direct} = \lambda_{S} \mathcal{L}_{S}
        +\lambda_{\Lambda} \mathcal{L}_{\Lambda}
        +\lambda_1 || \mathbf{\Lambda} ||_1
        +\lambda_2 || \mathbf{\Lambda}- \mathbf{\Xi}_{S}||_2^2 \label{eq:total_sindy_loss_direct}
\end{equation}
where the term $\mathcal{L}_{KAN}$ is removed, since there is no longer an underlying KAN to optimize.

Direct SINDy-KANs can be significantly faster than indirect SINDy-KANs because the method does not require solving a least squares problem at each activation function during each iteration. From Table \ref{tab:direct}, this corresponds to about a 33\% reduction in the computational cost for training for the example problem from Sec. \ref{sec:reg_test2}. However, in our tests, direct SINDy-KANs can be more challenging to train than SINDy-KANs as presented above. For this reason, we present some results with direct SINDy-KANs here, and leave further exploration of the stability and convergence of training direct SINDy-KANs for future work. 

\begin{table}[t]
    \centering
    \begin{tabular}{ l | l l  } 
     \hline\hline
       & $\mathcal{K}_\Lambda^{direct}$ & $\mathcal{K}_\Lambda$ (From Sec. \ref{sec:reg_test2})  \\     \hline\hline
    Exact eq. $f(x, y) = \cos(x^2 + y)$ & $0.99986\cos(0.99996x^2 + 0.99995y)$ & $0.99990\cos(1.00001x^2 + 0.99994y)$ \\ 
    Training time (s) & 145 & 215
    \end{tabular}
    \caption{Comparison of SINDy-KANs and direct SINDy-KANs.}
    \label{tab:direct}
\end{table}

We can also consider the ODE case from Sec. \ref{sec:ODE_2}, given by 
  
    \begin{equation}\frac{d}{dt}\begin{bmatrix} x \\ y \\ z\end{bmatrix}=
\mathbf{B}
    \begin{bmatrix} x \\ y \\ z \end{bmatrix}
    \end{equation}
    where
    \begin{equation}
        \mathbf{B} =     \begin{bmatrix}
        -0.1 & 2 & 0\\
        -2 & -0.1 & 0 \\
        0 & 0  & -0.3
    \end{bmatrix}.
    \end{equation}
From Table~\ref{tab:direct_ODE}, the SINDy-KANs and direct SINDy-KANs have comparable performance. 
    
\begin{table}[t]
    \centering
    \begin{tabular}{ l | c c  } 
     \hline\hline
       & $\mathcal{K}_\Lambda^{direct}$ & $\mathcal{K}_\Lambda$ (From Sec. \ref{sec:ODE_2})  \\     \hline\hline
$        \mathbf{B} =     \begin{bmatrix}
        -0.1 & 2 & 0\\
        -2 & -0.1 & 0 \\
        0 & 0  & -0.3
    \end{bmatrix}$ & $  \begin{bmatrix}
       -0.0989 & 1.9941 & 0\\
        -1.9967& - 0.0989 & 0 \\
        0 & 0  & -0.2964
    \end{bmatrix}$ & $       \begin{bmatrix}
        -0.0993 & 1.9946 & 0\\
        -1.9972 & - 0.0995 & 0 \\
        0 & 0  & -0.2811
    \end{bmatrix}$  
    \end{tabular}
    \caption{Comparison of SINDy-KANs and direct SINDy-KANs for the ODE problem in Sec. \ref{sec:ODE_2}.}
    \label{tab:direct_ODE}
\end{table}

\clearpage
\section{SINDy-KANs with noise} \label{app:noise}
\begin{figure}[t]
    \centering   
    \includegraphics[width=0.8\textwidth]{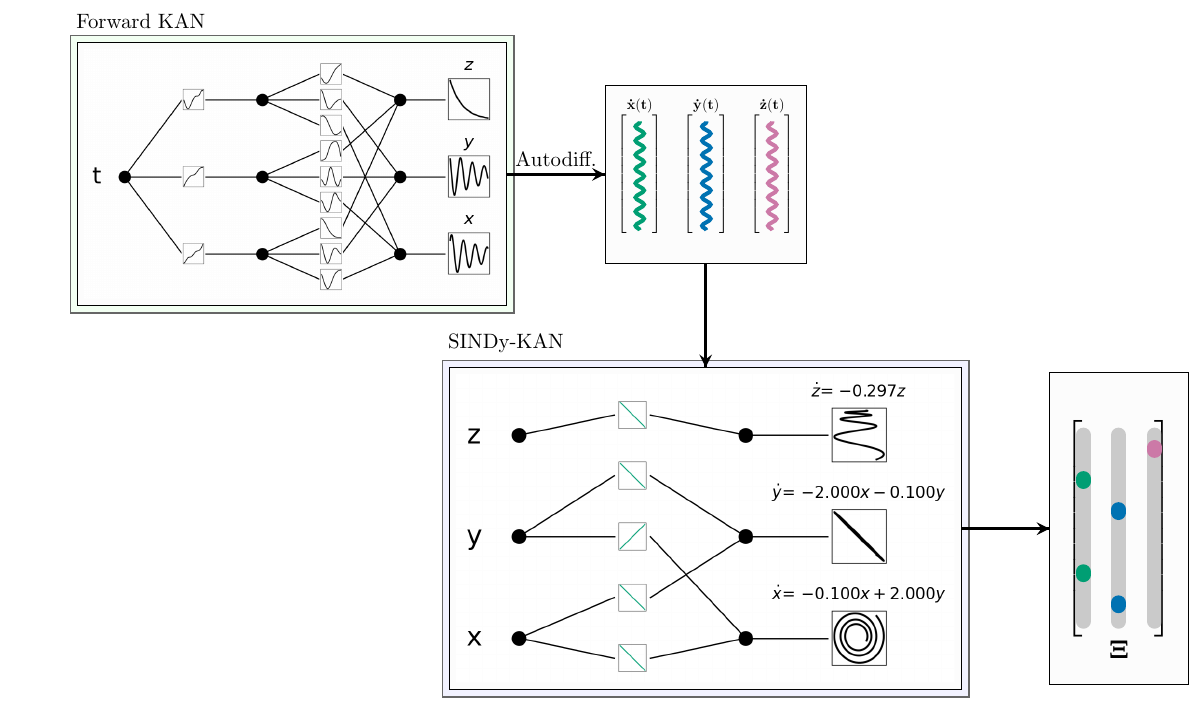}
    \caption{An illustration of SINDy-KANs for dynamical systems. When training with data, the derivatives can either come from a pre-trained ``forward KAN'' or be found through finite differences or other methods. }
    \label{fig:DS_diagram}
\end{figure}

We consider the extension of SINDy-KANs to noisy data in this section using the test case from~\cref{sec:ODE_2}. We progressively add more noise to the training data for $\mathbf{X}$, which increases the error in prediction  of $\mathcal{K}(\mathbf{X})$. The accuracy of the learned equation depends directly on the derivative. To take the derivative, we first train a KAN to learn the map $\mathbf{x} \rightarrow \mathbf{X}$, denoted by $\mathcal{K}(\mathbf{X})$ (we call this the ``forward KAN''). We then numerically take the derivative of the KAN prediction $\dot{\mathcal{K}}(\mathbf{X})$ using the autodifferentiation feature in \texttt{JAX}. An illustration of this process is given in Fig. \ref{fig:DS_diagram}. 

\begin{figure}[t]
    \centering    \includegraphics[width=\textwidth]{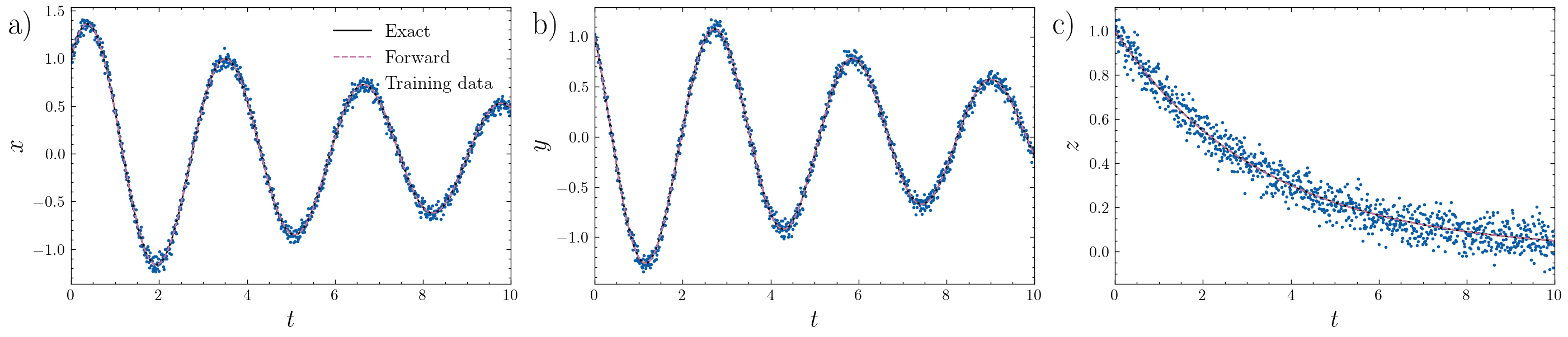}\\
     \includegraphics[width=\textwidth]{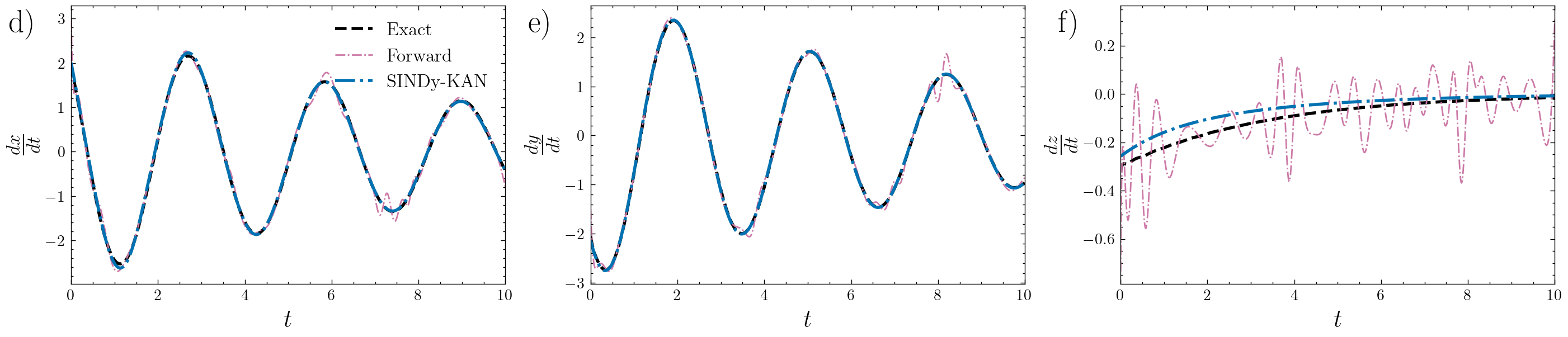}
\caption{Results for the test case from~\cref{sec:ODE_2} with 31.37\% relative noise added to the training data (a-c). Although the derivatives used to train the SINDy-KAN are noisy (d-f), the SINDy-KAN is able to recover most of the equation.}
    \label{fig:sindy_noise}
\end{figure}

We find that the SINDy-KANs are relatively robust up to 30\% relative noise for this problem. At high noise level, higher order terms ($(\cdot)^3$) are erroneously learned, which could be alleviated by restricting the space of candidate functions for the SINDy-KANs. Alternatively, other methods for approximating the derivative $\dot{\mathbf{X}}$ could be used as have been successfully pursued in previous work with SINDy. Results for the case with the most noise are given in~\cref{fig:sindy_noise}.

\begin{table}[t]
    \centering
    \begin{tabular}{c  c c c} 

     Data $\mathbf{X}$ relative noise &   $\mathcal{K}(\mathbf{X})$ error & $\dot{\mathcal{K}}(\mathbf{X})$ error & Learned equation\\ 
    \hline \hline    
    0\% & 5.688e-08 & -- & 
        \begin{tabular}{@{}l@{}}
    $\frac{dx}{dt} = -0.0997x + 2.00044y$ \\ 
    $\frac{dy}{dt} = -2.00017x - 0.09963y$ \\
    $\frac{dz}{dt} = -0.29567z$
            \end{tabular}
\\ \hline

     6.27\% & 1.255e-05 & 0.00014 & 
    \begin{tabular}{@{}l@{}}
    $\frac{dx}{dt} = -0.10575x + 1.99087y$ \\ 
    $\frac{dy}{dt} = -2.00280x - 0.10184y$ \\
    $\frac{dz}{dt} = -0.26376z$
            \end{tabular}
    \\\hline

     31.37\% & 0.00043 & 0.0101 & 
             \begin{tabular}{@{}l@{}}
    $\frac{dx}{dt} = -0.12326x + 0.10216y^3 + 1.93752y + 0.0897z^3$ \\ 
    $\frac{dy}{dt} = -2.00878x - 0.12078y$ \\
    $\frac{dz}{dt} = -0.09579z^3 - 0.16220z$
            \end{tabular}
    \end{tabular}
    \caption{Relative errors and learned equations for the ODE case from~\cref{sec:ODE_2} with noise added to the training data. Errors are computed by the relative mean squared error (MSE).}
    \label{tab:noise}
\end{table}

\clearpage
\section{Training parameters} \label{app:params}

\begin{table}[h]
    \centering
    \begin{tabular}{c  c } 
     \hline\hline
     Parameter &   Sec. \ref{sec:reg_test2} \\ 
     \hline\hline
    KAN architecture & [2, 2, 1]\\ 
    $g$       &        [3, 6]  \\ 
    Learning rate scales &   [1, 0.6]       \\
    Boundaries &           [0, 40k]          \\
    $k$   &            5     \\ 
    ADAM learning rate & 1e-3   \\ 
    ADAM iterations& 210k\\
    $\lambda_{S}$   & 1  \\
    $\lambda_{\Lambda}$ & 1 \\
    $\lambda_{1}$  & 0.0001  \\
    $\lambda_{2}$  & 0.0001 \\
    \hline
    \end{tabular}
    \caption{Hyperparameters used for training the results in~\cref{sec:reg_test2}. $g$ is the number of grid points used in the KANs. The grid is refined at the schedule denoted by the boundaries, and the learning rate is scaled at the same boundaries. $k$ is the polynomial degree for the B-splines.}
    \label{tab:params_regression}
\end{table}

\begin{table}[t]
    \centering
    \begin{tabular}{c  c  c c c c} 
     \hline\hline
     Parameter &  Sec. \ref{sec:ODE_2}  &  Sec. \ref{sec:ODE_3}   &  Sec. \ref{sec:ODE_5}   &  Sec. \ref{sec:ODE_6}   &  Sec. \ref{sec:ODE_7}  \\ 
     \hline\hline
    KAN architecture &      [3, 3]   &[2, 2] &    [3, 6, 3]  &  [3, 5, 3]  & [3, 6, 3]\\ 
    $g$   &                 [3]    &  [5, 10] &   [5, 7]     &  [5, 7, 10]  & [5, 7, 10, 15]\\ 
    Learning rate scales &  [1] &     [1, .6]  &   [1, .6]    & [1, .6, .6] & [1, .6, .6, .1]\\
    Boundaries &            [0] &     [0, 50k] & [0, 30k]    & [0, 30k, 60k] & [0, 30k, 60k, 300k]\\
    $k$   &                 5       &   5 & 5  &   5 & 5   \\ 
    ADAM learning rate &    1e-3    & 1e-3   &  5e-3    &  5e-3 & 1e-2 \\ 
    ADAM iterations &       20k   &  150k    & 50k       & 100k & 500k\\
    $\lambda_{S}$  &        1       & 1      & 10        & 10 & 0.1 \\
    $\lambda_{\Lambda}$ &   1       & 1      & 10        &100 & 10\\
    $\lambda_{1}$ &         0.0001  & 0.0001 & 0.001     &  0.001  &  0.0001 \\
    $\lambda_{2}$ &         0.001   & 0.001  & 0.001     & 0.001 &  0.0001 \\
    \hline

    \end{tabular}
    \caption{Hyperparameters used for training the results in~\cref{sec:ODE}. $g$ is the number of grid points used. The grid is refined at the schedule denoted by the boundaries, and the learning rate is scaled at the same boundaries. $k$ is the polynomial degree for the B-splines used. }
    \label{tab:params_ODE}
\end{table}

\end{document}